\documentclass{article} 
\usepackage{iclr2025_conference,times}

\usepackage{amsmath,amsfonts,bm}

\def\eqref#1{equation~\ref{#1}}

\def\1{\bm{1}}

\DeclareMathAlphabet{\mathsfit}{\encodingdefault}{\sfdefault}{m}{sl}
\SetMathAlphabet{\mathsfit}{bold}{\encodingdefault}{\sfdefault}{bx}{n}

\usepackage{hyperref}
\usepackage{url}
\usepackage{graphicx} 
\usepackage{booktabs}
\usepackage{subcaption}
\usepackage{algorithm}
\usepackage{algorithmicx}
\usepackage{algpseudocode}
\usepackage{wrapfig,lipsum}
\usepackage{tcolorbox}
\usepackage{verbatim}
\usepackage{amsmath}
\usepackage{amsfonts}
\usepackage{amssymb}
\usepackage{graphicx}
\usepackage{array}
\usepackage{listings}

\title{The KoLMogorov Test: \\  Compression by Code Generation}

\author{Ori Yoran$^{1,2}$\thanks{Work done during an internship at Meta FAIR.}, Kunhao Zheng$^{1}$, Fabian Gloeckle$^{1}$\\ 
\textbf{Jonas Gehring$^{1}$, Gabriel Synnaeve$^{1}$, Taco Cohen$^{1}$} \\
~$^1$Meta AI (FAIR), ~$^2$Tel Aviv University\\
\texttt{ori.yoran@cs.tau.ac.il} \\ \texttt{\{kunhao, fgloeckle, jgehring, gab, tscohen\}@meta.com} \\
}

\iclrfinalcopy 
\newif\ifcomments
\commentstrue

\ifcomments
    \newcommand\oy[1]{\textcolor{red}{[OY: #1]}}
    \newcommand\taco[1]{\textcolor{magenta}{[TC: #1]}}
    \newcommand\gab[1]{\textcolor{blue}{[GS: #1]}}
    \newcommand\kunhao[1]{\textcolor{purple}{[KH: #1]}}
\else
    \providecommand{\oy}[1]{}
    \providecommand{\taco}[1]{}
    \providecommand{\gab}[1]{}
    \providecommand{\kunhao}[1]{}
\fi

\newcommand\kt{\textsc{KoLMogorov-Test}}

\newcommand\codegenllm{\textsc{CodeLM}}
\newcommand\codegenllms{\textsc{CodeLMs}}
\newcommand\llamasmall{\textsc{Llama-3.1-8B}}
\newcommand\llamalarge{\textsc{Llama-3.1-70B}}
\newcommand\llamaxl{\textsc{Llama-3.1-405B}}
\newcommand\gpt{\textsc{GPT4-o}}
\newcommand\omini{\textsc{o1-mini}}
\newcommand\seqcodertiny{\textsc{SeqCoder-1.5B}}
\newcommand\seqcodersmall{\textsc{SeqCoder-8B}}
\newcommand\seqcoder{\textsc{SeqCoder}}
\newcommand\repeatseq{\textsc{Repeat}}

\newcommand\lmis{\textsc{LMiC}}
\newcommand\gzip{\textsc{Gzip}}

\renewcommand{\sfdefault}{cmss}
\newcommand\precision{\textsf{Precision}}
\newcommand\pass{\textsf{Accuracy}}
\newcommand\compressionrate{\textsf{CompressionRate}}

\begin{document}

\maketitle

\begin{abstract}
Compression is at the heart of intelligence. A theoretically optimal way to compress any sequence of data is to find the shortest program that outputs that sequence and then halts.
However, such \emph{Kolmogorov compression} is uncomputable, and code generating LLMs struggle to approximate this theoretical ideal, as it requires reasoning, planning and search capabilities beyond those of current models.
In this work, we introduce the \kt{} (KT), a compression-as-intelligence test for code generating LLMs. In KT a model is presented with a sequence of data at inference time, and asked to generate the shortest program that produces the sequence. We identify several benefits of KT for both evaluation and training: an essentially infinite number of problem instances of varying difficulty is readily available, strong baselines already exist, the evaluation metric (compression) cannot be gamed, and pretraining data contamination is highly unlikely.
To evaluate current models, we use audio, text, and DNA data, as well as sequences produced by random synthetic programs.
Current flagship models perform poorly -- both \gpt{} and \llamaxl{} struggle on our natural and synthetic sequences. On our synthetic distribution, we are able to train code generation models with lower compression rates than previous approaches. Moreover, we show that gains on synthetic data generalize poorly to real data, suggesting that new innovations are necessary for additional gains on KT.

\end{abstract}

\section{Introduction}
\begin{figure}[t]
  \includegraphics[width=1.0\columnwidth]{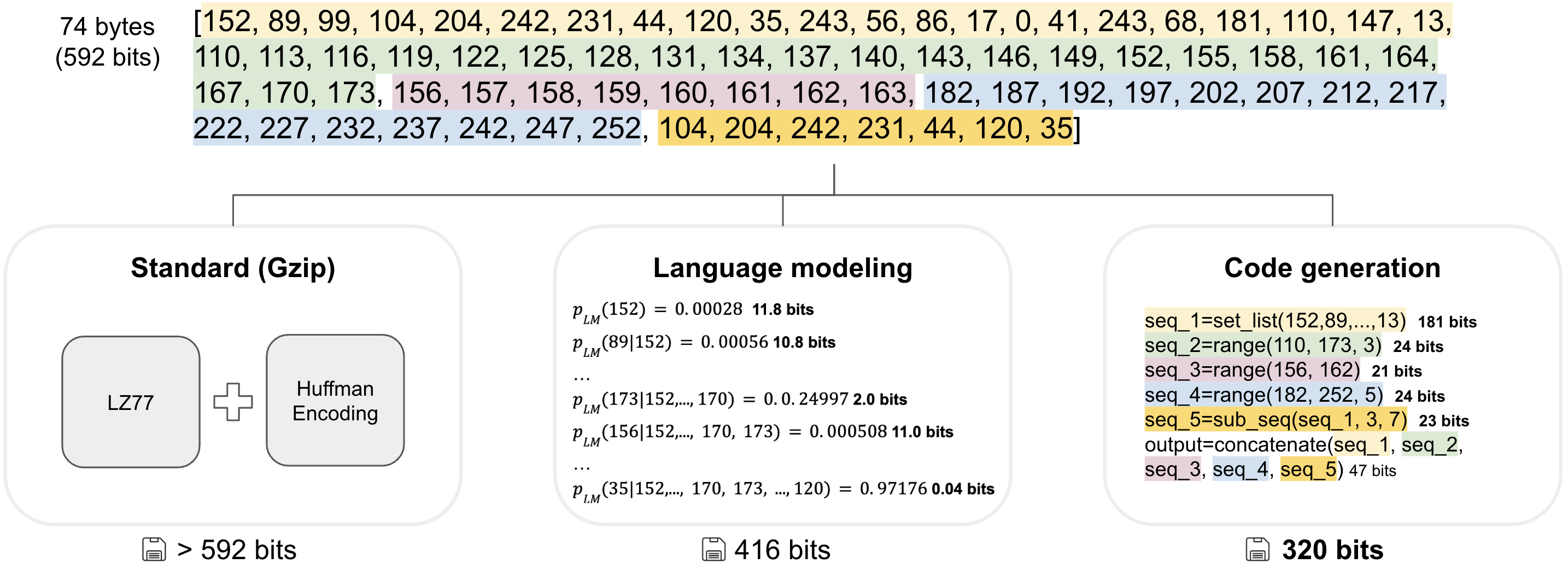}
  \caption{\textbf{Data compression by code generation.} Consider compressing a sequence of bytes (presented as numbers in range $[0, 255]$) that can be produced by composing simpler sub-sequences. Standard compression methods, such as \gzip{}, focous on repetitions and frequency of characters and fail to exploit the logical patterns in this sequence (although they are strong baselines for long sequences, \S\ref{subsec:analysis}). LLMs are better at finding complex patterns, such as a sequence of incremental numbers, and can be used for compression with arithmetic coding. However, they are sensitive to phase-shifts  due to their auto-regressive manner, and require model weights for decoding. Code generative models, inspired by the concept of Kolmogorov Complexity, can identify patterns in the input sequence to generate concise programs whose execution produces the original sequence.}
  ~\label{fig:fig-1}
  \vspace{-10pt}
\end{figure}

Compression and code generation are deeply related through the notion of Kolmogorov complexity, denoted $K(x)$, which is defined as the length of the shortest computer program\footnote{For some universal Turing machine $U$.} that produces the sequence $x$ as output and hence constitutes the optimal compression of $x$ \citep{b3dc75af-1a78-39c9-b837-c7f3ee48d5d5, book, Hutter2024-mb} (see \S\ref{sec:background} for a detailed background). Kolmogorov complexity is \emph{uncomputable} as it reduces to the halting problem, making the search for improved computable \emph{upper bounds} a never-ending challenge and a potential benchmark for intelligence that by definition cannot saturate.
We propose to view code generation language models (\codegenllms{}) as upper bounds for Kolmogorov complexity: we task them to identify patterns in an input sequence and to compress it by producing a short program that outputs said sequence, and can measure their accuracy at producing correct programs as well as the compression rates achieved. This is the \kt{} (KT), a benchmark for reasoning capabilities for \codegenllms{}, illustrated in Fig.~\ref{fig:fig-1}.

We point out the following benefits of compression by code generation as a benchmark for the reasoning capabilities of language models:
(1) the compression metric can be trusted in that it does not produce false positives,
(2) diverse and richly structured sequence data is abundantly available,
(3) it is highly unlikely that pretrained models have seen many relevant (program, sequence) pairs, making memorization-based solutions infeasible\footnote{Models trained on synthetic data or by RL could very well end up using memorization.},
(4) if the benchmark is saturated, either through improved reasoning and search capabilities or memorization, we can simply increase the sequence length and
(5) following research spanning decades of theoretical computer science, classical compression algorithms such as \gzip{} \citep{10.17487/RFC1952} can serve as strong baselines.

To evaluate current \codegenllms{} on KT, we use naturally occurring sequences from three data modalities: text, audio, and DNA (\S\ref{subsec:real_data}). As we do not know the optimal program for the sequences, it is not possible to use this data for supervised learning, making it ideal for evaluation.
To collect program-sequence pairs for supervised training and evaluation, we design a \emph{compositional} domain-specific language (DSL) coupled with an automatic data generation framework (\S\ref{subsec:syn_data}), inspired by context-free grammars \citep{1056813, 10.5555/1196416}.
 
Our experiments (\S\ref{sec:exps}) show that sequence compression is an extremely challenging task for current \codegenllms{}. Strong \llamaxl{} \citep{dubey2024llama3herdmodels} and \gpt{} \citep{openai2024gpt4technicalreport} models generate Python programs that fail to produce the input sequence in $78\%$ and $40\%$ of the time for naturally occurring data, and $66\%$ and $45\%$ percent of the time for synthetic sequences that follow clear patterns. On our synthetic distribution, we are able to train relatively small \codegenllms{} with 1.5B parameters that outperform state-of-the-art prompted models by more than $40\%$, but struggle on real sequences. We further find that the prior used for program-compression is an important factor in overall compression performance, with a simple uniform prior over our DSL outperforming \gzip{}.
Finally, we observe modest gains for adding inline execution feedback.

In \S\ref{subsec:analysis}, we conduct a thorough analysis of our results.  We find that gains on synthetic data partially generalize to real data -- models trained for longer perform better on short sequences, but all current models perform poorly on long sequences, suggesting KT can act as a useful test-bed to evaluate scaling properties of new methods.
Finally, we perform an error analysis and find that models make a wide range of errors, from generating over-simplified programs that do not produce the input sequence, to repeating the input sequence in over-complicated ways.

To summarize, our main contributions are:\footnote{To support future progress, we release our code and data at \url{https://github.com/facebookresearch/kolmogorov} and a public leaderboard at \url{https://huggingface.co/spaces/KoLMogorov-Test/Leaderboard}.}

\begin{itemize}
    \item We propose the \kt{}, an extremely challenging compression-as-intelligence test for \codegenllms{} (\S\ref{sec:data}).
    \item We show that \codegenllms{} can outperform previous compression methods on synthetic distributions where sampling program-sequence pairs is possible and efficient priors can be used, but fare very poorly on real data (\S\ref{sec:results}).
    \item We show that performance on real data scales poorly with increasing synthetic dataset size, suggesting that new breakthroughs may be needed for further progress on KT.
\end{itemize}
\section{Background}
\label{sec:background}

\paragraph{Generative Modelling, Information Theory and Compression.}

Generative modelling and compression are deeply related.
Given a generative model over sequences (e.g., autoregressive transformer \citep{radford2018improving}) $p(x_i | x_{<i})$, one can use the arithmetic coding algorithm to compress a particular sequence $x$ to a bitstream of length about $-\sum_i \log_2 p(x_i | x_{<i})$ bits \citep{5391119, 1055739}.
If the sequences are sampled from the gold distribution $p^*$, then the expected arithmetic code length is the cross entropy between $p^*$ and $p$.
When $p=p^*$, the cross entropy equals the entropy, which is the fundamental limit on average compression length for data sampled from $p^*$.

Similarly, given a latent variable model $p(x, z)$, one can encode a sequence $x$ by first encoding a code $z$ using \emph{prior} $p(z)$ and then encoding $x$ using likelihood $p(x|z)$.
Using an optimal code for $p(z)$ and $p(x|z)$ the coding cost will be roughly $-\log p(z) - \log p(x|z)$.
If we obtain $z$ by sampling from an encoder network $q(z|x)$, the expected coding cost will be $\mathbb{E}_q[-\log p(z) - \log p(x|z)]$ \citep{Habibian2019}, which (up to an additional entropy bonus $H(q)$) equals the evidence lower bound (ELBO) used for instance for VAE training \citep{kingma2022autoencodingvariationalbayes}.
Thus, both for autoregressive and latent variable models, maximizing likelihood is maximizing compression.

\paragraph{Algorithmic Information Theory, Solomonoff Induction, and Kolmogorov Complexity.}

In classical generative modelling and compression, one is concerned with finding a generative model from a particular model class that can be used to compress data from a particular distribution.
By contrast, in algorithmic information theory, one considers as ``model class'' the set of all computable functions.
This class trivially contains the optimal compression, and changing the computational model (programming language / universal Turing machine) only incurs a constant overhead depending on the pair of languages involved and regardless of the input \citep[2.2.2]{grunwald2008algorithmicinformationtheory}.

In analogy to the discussion above, we can sketch the theory as follows, inspired by Solomonoff's theory of inductive inference \citep{SOLOMONOFF19641}.
We consider the \emph{universal prior} $p(\rho) = 2^{-l(\rho)}$ over programs $\rho$ (where $l(\rho)$ denotes the length) that assigns higher probability to shorter programs (``Occam's Razor'').
Furthermore, we consider the likelihood $p(x | \rho)$ over output sequences $x$ given $\rho$ which puts all mass on the actual output of $\rho$.

Then, treating $\rho$ as latent we may encode $x$ using a two-part code by finding a \emph{program} $\rho$ that outputs $x$ (e.g. using a neural network $q(\rho|x)$), and encoding it using the \emph{prior} $p(\rho)$.
As before, the coding cost will be $\mathbb{E}_q[-\log p(\rho) - \log p(x|\rho)]$, where the second term is $\infty$ whenever $x$ is not the output of $\rho$, and $0$ when it is.
Clearly, the optimal $\rho$ is the shortest program that outputs $x$.
The length of this program $\rho^*$ is called the \emph{Kolmogorov Complexity} $K(x)$ \citep{b3dc75af-1a78-39c9-b837-c7f3ee48d5d5, book}.

\section{Data Collection}
\label{sec:data}

\begin{figure}[t]
\includegraphics[width=1.0\columnwidth]{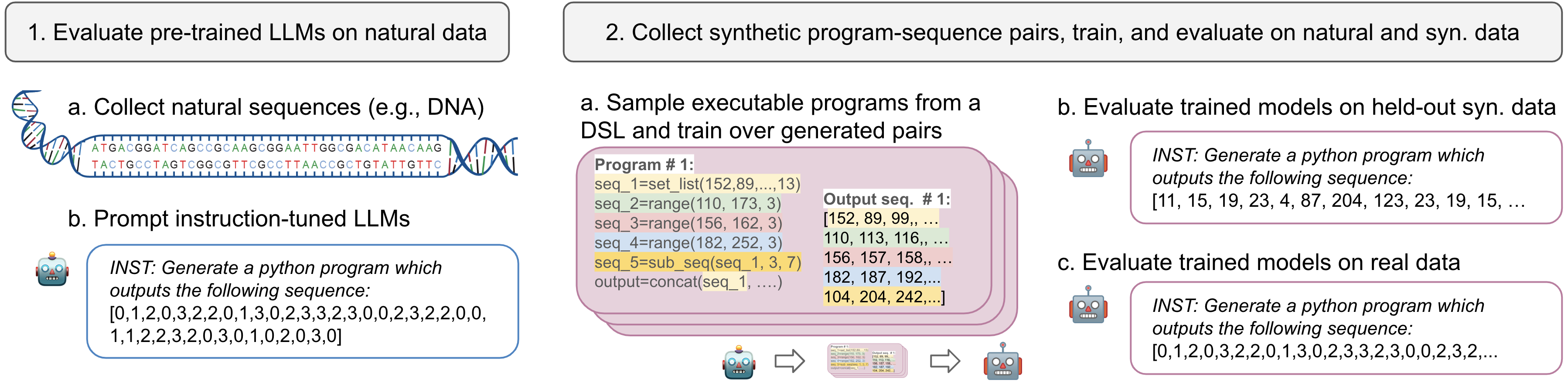}
  \caption{\textbf{Our main experimental settings.}}
  ~\label{fig:fig2}
  \vspace{-10pt}
\end{figure}

\subsection{Problem Setting}
The theory presented above suggests an interesting challenge for \codegenllms{}: to generate concise \emph{programs} (under well-defined \emph{priors}) that output a given sequence (Fig.~\ref{fig:fig-1} and Alg.~\ref{alg:kt} in \S\ref{subsec:kt_alg}). 
We evaluate \codegenllms{} on three naturally occurring modalities: audio, text, and DNA (Fig.~\ref{fig:fig2}, left, and \S\ref{subsec:real_data}).
We focus on audio and textual data, popular modalities in compression works with strong baselines \citep{deletang2024language}. We also introduce DNA sequences as these follow simple biological patterns.
Although real data is useful for evaluation since it can be obtained in large amounts for many modalities, corresponding \emph{programs} required for supervised training are not available.
As a remedy, we experiment with synthetic settings where we collect program-sequence pairs by sampling programs from a Domain Specific Language (DSL) and executing the sampled programs (Fig.~\ref{fig:fig2}, right, and \S\ref{subsec:syn_data}).

\subsection{Naturally Occurring Sequences}
\label{subsec:real_data}

In all our experiments on natural data, the model has to compress 1MB of information. The length of the input sequence presented to our models ranges from 16 to 1024 bytes depending on the setting.

\paragraph{Audio.} We randomly sample audio snippets from the LibriSpeech development and test sets \citep{7178964}. We parse the data to three audio formats: (a) high quality audio with 16 bits depth, from which the original data can be perfectly recreated but each sample is split across two bytes, (b) lower quality audio with 8 bits depth, which is simpler for our baselines (\S\ref{sec:results}), and (c) an estimation of the Mel-Frequency Cepstral Coefficients encoding (MFCC) which does not allow direct reconstruction but describes the main features of the sound (see \S\ref{appendix:natural} for more details). Each byte is represented as a number in range $[0, 255]$.

\paragraph{Text.}
Following recent work \citep{deletang2024language}, we use the enwik9 Wikipedia corpus \citep{hutter2009hutter}. The Unicode text is encoded to a sequence of bytes with UTF-8 encoding which are represented as a list of numbers.

\paragraph{DNA.}
We use Genome assembly GRCh38 which contains $3.1$GB of human DNA in FASTA format \citep{grch38}. Since the original files include upper and lower case variants of each of the four nucleotides, we use a vocabulary of eight characters represented by numbers in range $[0, 7]$\footnote{Lower case variants represent low-complexity or masked regions (e.g., repeats or predicted sequences), which we keep to allow perfect reconstruction of the input sequence.}.

\subsection{Synthetic Data Generation}
\label{subsec:syn_data}

Our aim is to examine whether \codegenllms{} can be trained to generate concise and correct programs. Hence, we adhere to the following desiderata for our data generation process: (a) \textbf{completeness}: each sequence has some probability for a program that produces it to be sampled,
and (b) \textbf{simplicity}: 
shorter programs will be sampled with higher probability.

\paragraph{Domain Specific Language.}
We design a \emph{compositional} DSL, in the sense that the output sequence is created by \emph{composing} simpler sub-sequences (Fig.~\ref{fig:fig2}, center). Our DSL supports the following four classes of functions (see \S\ref{appendix:syn_data} for a list of all supported functions):\footnote{We note that for simplicity, our DSL does not allow recursion -- it is a proper subset of primitive recursive functions and of Turing-computable functions \citep{10.1112/plms/s2-42.1.230} and thus deviates from the theory in \S\ref{sec:background}.}
\begin{itemize}
    \item \textbf{Sequence initiators}: functions that take variables as input and return a sequence of numbers. These include a range of increasing numbers, a repetition of a single number, or a fixed (hard-coded) list of numbers.
     \item \textbf{Sequence modifiers}: functions that take a sequence as input and return a modified version of the sequence, e.g., by reversing, repeating, or substituting elements from the sequence. A sub-class of our modifiers includes mathematical operations, such as scan-adding the elements in the sequence or applying a modulo operation.
     \item \textbf{Sequence filters}: functions that filter a sequence, e.g., by keeping only even values.
    \item \textbf{Sequence mergers}: functions that take two sequences and merge them to a single one: concatenation, interleaving, pointwise addition, subtraction or modulo of the sequences.
    
\end{itemize}

\begin{figure}[t]
\includegraphics[width=1.0\columnwidth]{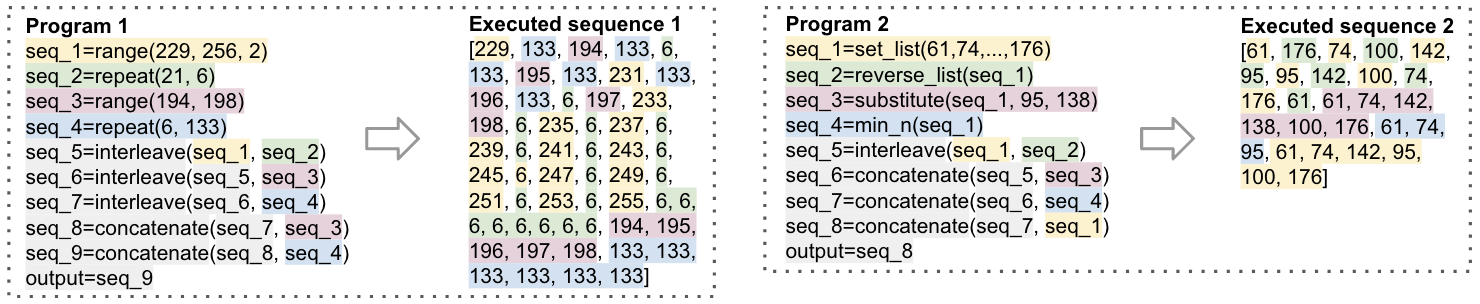}
  \caption{\textbf{Two examples of program-sequence pairs from our synthetic data generation process.}}
  ~\label{fig:syn_examples}
  \vspace{-10pt}
\end{figure}

\paragraph{Sampling programs.}
Given the compositional nature of our DSL, we sample programs similarly to the way sequences are sampled from a Context Free Grammar \citep{1056813, 10.5555/1196416}. We can also control the distribution of programs by assigning priors over the program distribution, e.g., the lengths of the initiated sequence or probability to apply modifications. For simplicity and to avoid biases we keep the priors uniform\footnote{We note that this does not guarantee that each operator is used with the same probability, as some operators are only applicable in specific contexts (e.g., an addition between two sequences is only applicable when the sequences have the same length and the sum of the matching elements is in the allowed range).}. For examples of program-sequence pairs, see Fig.~\ref{fig:syn_examples}. 
As in the natural data domains, we generate 1MB of evaluation data. For training, we sample 1M pairs from the same distribution. We provide additional details and statistics in \S\ref{appendix:syn_data}.

\paragraph{Encoding programs.}
We encode programs written in our DSL using a factorized uniform prior over functions and arguments.
As each line includes a call to a single function, the encoding cost of each line is the cost of encoding the function (i.e., $\log_2(|functions|)$) plus the cost of the input parameters (e.g., the number of bits needed to set an arbitrary list for \emph{set list}, or the indices of the sequences in \emph{concatenate}).
Our code length calculation algorithm is presented in \S\ref{appendix:syn_data}, Alg.~\ref{alg:uni-encoding}.

\section{Experimental Setup}
\label{sec:exps}

\paragraph{Baselines.}
For our classical compression baseline, we use \gzip{} \citep{10.17487/RFC1952}.
We also include a Language Modeling is Compression (\lmis{}) baseline (see \S\ref{appendix:models} for implementation details), where the LM predictions are used together with arithmetic coding to compress the sequence as discussed in \S\ref{sec:background} \citep{deletang2024language}.
A major limitation of \lmis{} is that the original LLM is needed to decompress the data\footnote{As our focus is on compressing relatively small amounts of data, we report raw \lmis{} results disregarding the weights of the model, an upper bound of the true performance which requires the weights that were used for encoding at decoding time.}.  
We also include \repeatseq{}, a naive baseline that always returns the input sequence, and an \textsc{Upper Bound} baseline that returns the matching program for our synthetic pairs.

\paragraph{Zero-shot prompted baselines.}
We prompt our models to generate the shortest Python program that produces the input sequence (see \S\ref{appendix:models}, Fig.~\ref{fig:prompt} for the full prompt). We use strong open-weight models from the \textsc{Llama-3.1} family with 8, 70, and 405 billion parameters \cite{dubey2024llama3herdmodels}. In addition, we use \gpt{} as a closed-source alternative \citep{openai2024gpt4technicalreport}. 
We also experiment with a chain-of-thought (CoT) \citep{wei2022chain} baseline (see prompt in Fig.~\ref{fig:prompt-cot}) and with \textsc{OpenAI o1-mini} \citep{openai2024openaio1card}, a closed-source model trained to think before generating the final answer.
We use \gzip{} to encode programs before measuring their bit length (i.e., we use \gzip{} as a \emph{prior} over programs) in all experiments, unless stated otherwise.

\paragraph{Trained models.}
For our trained models, we use \llamasmall{} as base model in addition to a 1.5-billion parameter LLM with the same architecture, which we train on a mixture of open-source code and text (see \S\ref{appendix:models} for more details and technical specifications). We further train our models on 10K, 100K or 1M unique programs-sequence pairs sampled from our data generator (\S\ref{subsec:syn_data}), and name them \seqcoder{-10K}, \seqcoder{-100K} and \seqcoder{(-1M)}, respectively.
We use the encoding algorithm defined in \S\ref{subsec:syn_data} for encoding programs from our synthetic DSL.
 
\paragraph{Sequence lengths.}
We use a sequence length of $1024$ bytes for our \lmis{} baseline, based on findings that further increasing context length does not result in significant gains \citep{deletang2024language}. For natural data, we use a sequence length of $128$ bytes (see \S\ref{subsec:analysis} for a discussion on input lengths). Our synthetic data has an average sequence length of 75.9 bytes with a standard deviation of 73.7 (see \S\ref{appendix:syn_data}, Fig.~\ref{fig:syn_seq_length_hist} and Fig.~\ref{fig:program_length} for histograms  of sequence and programs lengths). 

\paragraph{Metrics.}

Given a sequence $x$ and a program candidate $\rho$ suggested by the model with output $[\rho]$, we define the variable $y$ that backs off to the \repeatseq{} baseline when $\rho$ errs by:
\[
y =
\begin{cases}
\rho \qquad \text{if}\; [\rho] = x, \\
x \qquad \text{otherwise}.
\end{cases}
\]
We define the following metrics:

\begin{align*}
\pass{}(x, \rho) &= \mathbf{1}_{\{[\rho] = x\}}, \\
\compressionrate{}(x, \rho) &= \frac{1+\|y\|_\mathrm{enc}}{\|x\|}, \\
\precision{}(x, \rho) &= \frac{\|\rho\|_\mathrm{enc}}{\|x\|_\mathrm{gzip}} \qquad \text{where}\; [\rho] = x,
\end{align*}
where $\|\cdot\|$ denotes the bit-length of the data, $\|\cdot\|_\mathrm{gzip}$ denotes the length after \gzip{}-encoding, and $\|\cdot\|_\mathrm{enc}$ denotes the length after encoding programs with \gzip{} for prompted models and our uniform encoding for \seqcoder{} models.
In other words, \pass{} measures whether the program is correct, \compressionrate{} measures the compression of the input when storing the compressed representation (an additional bit is saved to determine whether $y$ is $\rho$ or $x$ and lower \compressionrate{} is better\footnote{\compressionrate{} is higher than $1$ when the compressed representation is larger than the original.}), and \precision{} measures the compression of correct programs in relation to the \gzip{} baseline. \repeatseq{} has an \pass{}, \compressionrate{} and \precision{} of 1 by definition.

\section{Results}
\label{sec:results}

Running prompted pretrained models as well as smaller trained models on \kt{} with natural as well as synthetic sequence data sources, we make the following main findings: Zero-shot compression by code generation is a challenging benchmark even for the most powerful models available (\S\ref{subsec:kt_hard}). On short sequences of synthetic data, trained specialized models outperform the in-context reasoning baseline \lmis{} as well as classical baseline algorithms (\S\ref{subsec:synth_res}). In additional analyses, we examine the effect of scaling model and dataset size for models trained on synthetic data, their length generalization as well as typical failure modes (\S\ref{subsec:analysis}).

\subsection{\kt{} is Challenging for Current Models}
\label{subsec:kt_hard}

\begin{table*}[t]
\centering
\tiny
  \begin{tabular}{l|cc|cc|cc|cc|cc|cc}
    \toprule
    & \multicolumn{2}{c}{Audio-16-bit} & \multicolumn{2}{c}{Audio-8-bit} & \multicolumn{2}{c}{Audio-MFCC} & \multicolumn{2}{c}{DNA} & \multicolumn{2}{c}{Text}  & \multicolumn{2}{c}{Syn.} \\ 
    & 
    $\textsf{Acc.}$ & $\textsf{Prec.}$ &  $\textsf{Acc.}$ & $\textsf{Prec.}$ &  $\textsf{Acc.}$ & $\textsf{Prec.}$ &  $\textsf{Acc.}$ & $\textsf{Prec.}$ &  $\textsf{Acc.}$ & $\textsf{Prec.}$ &  $\textsf{Acc.}$ & $\textsf{Prec.}$\\ 
    \midrule
    \llamasmall& 5.9 &  1.54 & 3.9  & 1.74 & 8.8 & 1.51 & 3.7 & 2.34 & 1.4 & 3.12 & 8.5 & 2.48 \\
    \llamalarge & 18.0 & 1.96 & 10.1 & 1.67 & 24.2 & 1.56 & 22.5 & 2.18 & 9.6 & 3.17 & 18.0 & 2.78 \\
    \llamaxl & 35.6 & 1.66 & 15.0 & 1.66 & 29.6 & 1.58 & 24.8 & 2.06 & 6.5 & 3.17 & 33.3 & 2.18 \\ \midrule
    \gpt & \textbf{69.5} & \textbf{1.34} & \textbf{36.4} & \textbf{1.43} & \textbf{83.8} & \textbf{1.33} & \textbf{50.3} & \textbf{1.73} & \textbf{54.2} & \textbf{1.94} & 44.7 & 1.65 \\ \midrule
    \seqcodertiny &  0.0  & n/a &  $<$0.1  & n/a & 0.1  & n/a &  0.0 & n/a &  0.0  & n/a & 84.5 & 0.57 \\ 
    \seqcodersmall{} &  $<$0.1 & n/a & 0.1 & n/a &  0.1  & n/a  & 0.0 & n/a & 0.0 & n/a & \textbf{92.5} & \textbf{0.56} \\ 
    \bottomrule
    
\end{tabular}
\caption{\textbf{\pass{} and \precision{} for zero-shot prompted models on the \kt{} for sequences of length $128$.} Larger  models have higher \pass{}. On synthetic distributions when sampling program-sequence pairs is possible, our trained \seqcoder{} performs best and is the only model to achieve \precision{} that is lower than $1.0$, i.e., it outperforms \gzip{} when programs are correct. However, it performs poorly on real data.}
\label{tab:results_zero_shot}
\end{table*}

Tab.~\ref{tab:results_zero_shot} presents results for our prompted and \seqcoder{} models on different data modalities. Prompted models perform poorly -- strong \llamaxl{} and \gpt{} models err on average more than $78\%$ and $40\%$ of the time for naturally occurring sequences, with high variance between modalities, and more than $65\%$ and $55\%$ of the time on our synthetic data, which follows simple compositional rules by design.

Nevertheless, larger models perform better, with \llamaxl{} reaching the best \pass{} between \textsc{Llama} models, and \gpt{} reaching the highest \pass{} and lowest \precision{} overall.
All prompted models have a \precision{} score that is higher than $1.0$, meaning that on average, generated programs are larger than the \gzip{}-encoded sequences. We partly attribute this to \gzip{} being unoptimized to the generated Python programs, a phenomenon we discuss in more detail in \S\ref{subsec:analysis}. 

Our trained \seqcoder{} models significantly outperform all \codegenllms{} on synthetic data, and are the only models with a better \precision{} than the naive \repeatseq{} baseline (which has a perfect \pass{} and a \precision{} of $1.0$ by definition), suggesting that training data and strong priors are necessary for good performance. 
Nevertheless, it reaches near-zero performance on natural sequences. We discuss generalization to real data in more detail in \S\ref{subsec:synth_res} and \S\ref{subsec:analysis}.

We present results for \textsc{Llama} with CoT prompting in \S\ref{appendix:results}, Tab.~\ref{tab:results_zero_shot_cot}. We observe similar trends, albeit lower performance for \llamasmall{} and \llamalarge{} (a drop of $1.3$ and $5.1$ points in \pass{} on average across modalities), and a small increase of $2.1$ points in \pass{} for \llamaxl{}. 
For \omini{}, we observe overall similar results to \gpt{} (see \S\ref{appendix:results}, Tab.~\ref{tab:gpts} for mode details). 
Overall, our results are in line with recent work showing that CoT does not consistently improve performance for code generation tasks \citep{sprague2024cotcotchainofthoughthelps, zheng2024makeslargelanguagemodels}.

\begin{wrapfigure}{r}{0.35\textwidth}
  \centering
     \vspace{-3mm}

  \includegraphics[width=0.35\textwidth]
  {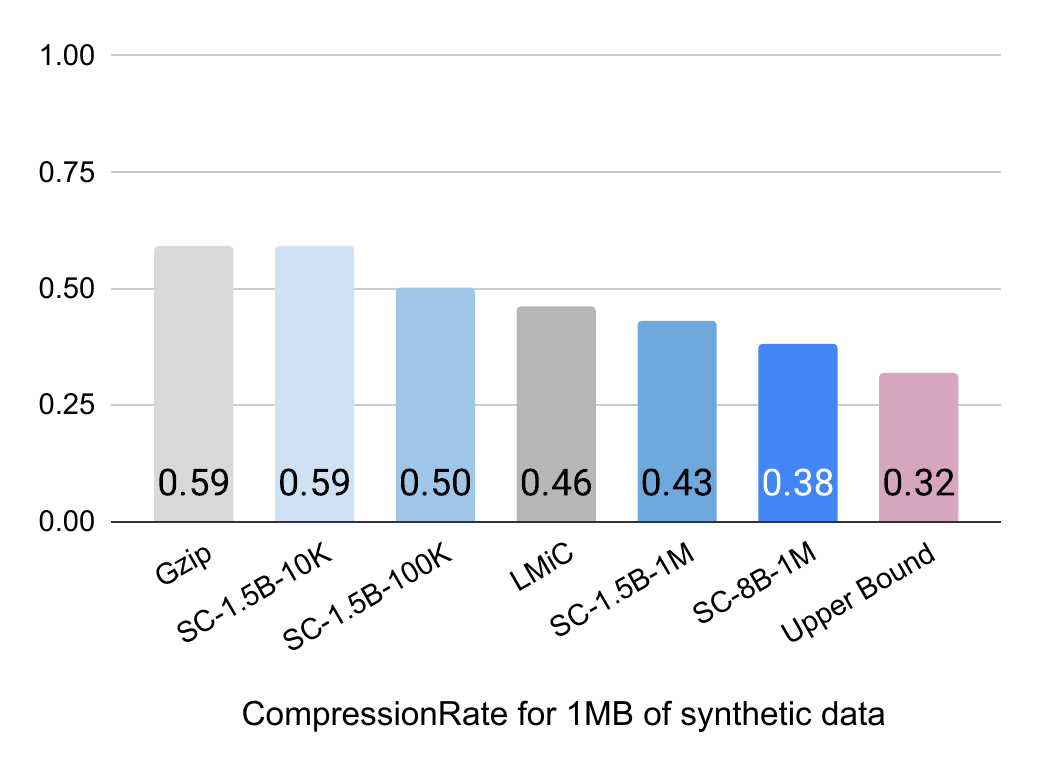}
  \caption{\textbf{\compressionrate{} for \seqcoder{} models trained on 10K-1M program-sequence pairs.} Models trained on enough data outperform the baselines.}

  \label{fig:sub1}
\end{wrapfigure}

\subsection{Trained \codegenllms{} Outperform Previous Compression Algorithms}
\label{subsec:synth_res}
Fig.~\ref{fig:sub1} presents \compressionrate{} on synthetic data for \seqcoder{} models. Even though instruction-tuned models struggle on this distribution, models can be trained to outperform previous compression algorithms, reaching a \compressionrate{} of $0.38$ for \seqcodersmall{}, outperforming the strong \gzip{} and \lmis{} baselines \emph{without requiring gigabytes of LLM weights} for decoding (in \S\ref{appendix:results} we empirically show that \gzip{} and \lmis{} are strong baselines and that scaling also improves \pass{} on synthetic data).

\begin{figure}[ht]
  \centering
  \begin{minipage}{0.45\textwidth}
    \centering
    \includegraphics[width=\textwidth]{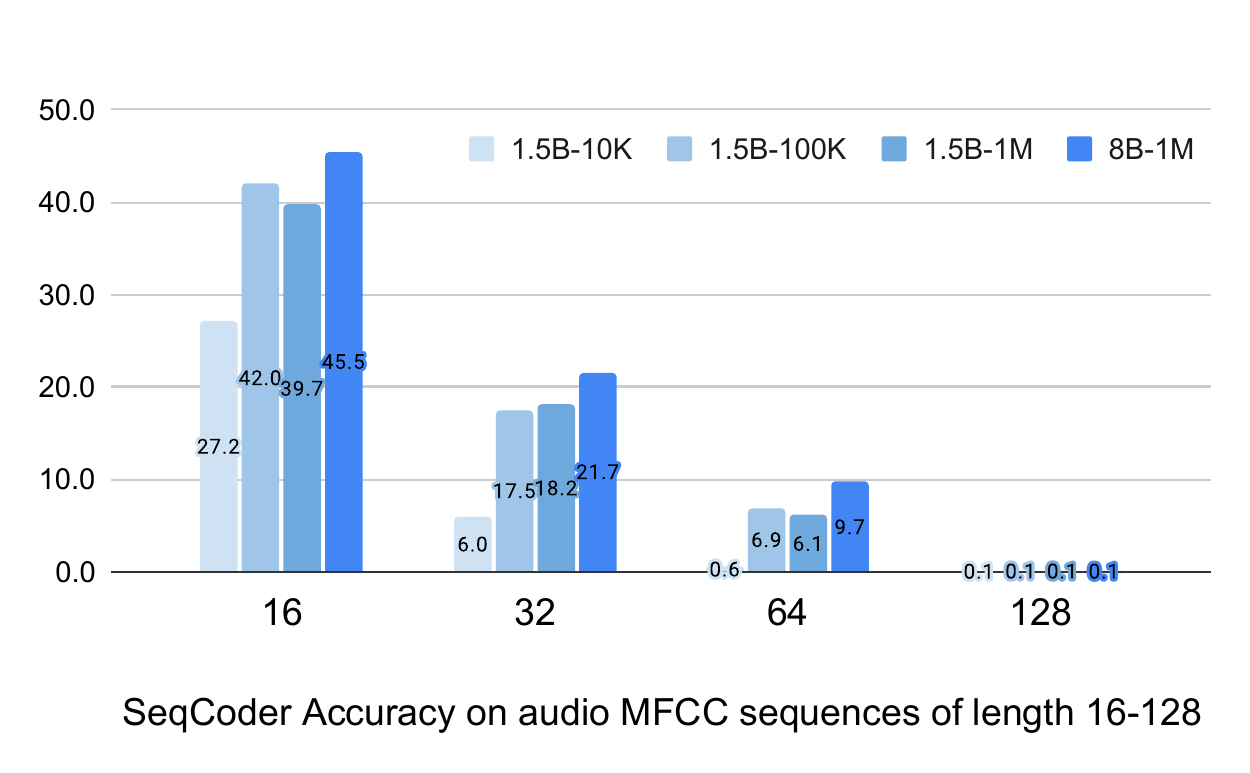}
    \caption{\textbf{\pass{} for our \seqcodertiny{} on Audio MFCC sequences of lengths 16-128.} Models trained on more examples are significantly better on short sequences, but all models struggle on longer ones.}
    \label{fig:sub2}
  \end{minipage}%
  \hfill
  \begin{minipage}{0.45\textwidth}
    \centering
    \includegraphics[width=\textwidth]{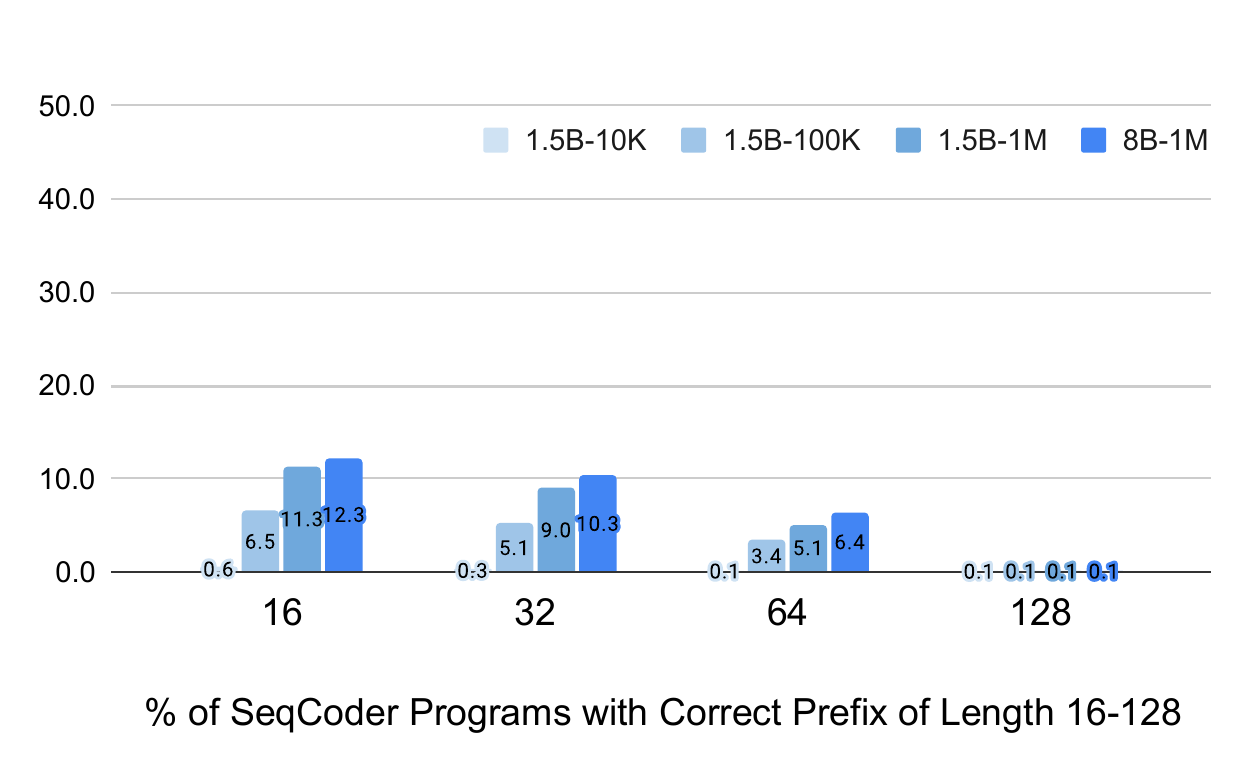}
    \caption{\textbf{The longest correct prefix length for \seqcoder{} on Audio MFCC sequences of length 128.} Models trained on more examples and more parameters generate longer correct prefixes.}
    \label{fig:prefix_length}
  \end{minipage}
  
\end{figure}

\paragraph{Length generalization.}
To better test if improvements on synthetic data generalize to real data, we evaluate \seqcoder{} models on 10,000 shorter Audio-MFCC sequences of lengths 16-64, presented in Fig.~\ref{fig:sub2}. All models perform poorly on longer sequences of length 64. Models trained on 100K and 1M examples significantly outperform \seqcodertiny{-10K} on shorter sequences, and larger \seqcodersmall{} perform best.

Fig.~\ref{fig:prefix_length} presents the length of correct prefixes for \seqcoder{} on Audio-MFCC sequences of length 128. \seqcodertiny{} models trained on 10K examples fail early, i.e., more than $99\%$ of programs have an error in the first 16 characters, even though the model correctly produces programs for $27.2\%$ of sequences of the same length (Fig.~\ref{fig:sub2}). Models trained on 100K and 1M examples perform better, but they exhibit similar trends, and err in the first 16 characters in $93.5\%$ and $88.7$\% of test cases.
We provide additional evidence that 
longer synthetic programs and sequences are challenging for \seqcodertiny{} in \S\ref{appendix:analysis}, Fig.~\ref{fig:syn_program_lenghts} and Fig.~\ref{fig:syn_sequence_lenghts}.

Tab.~\ref{tab:length_audio} presents results for \gzip{} and \llamaxl{} on a random sample of 2,000 Audio-8-bit sequences of lengths 16-128. As expected, \compressionrate{} for \gzip{} improves with length, as it is able to better exploit patterns in the input data (we extend to longer sequences for \gzip{} and \lmis{} in \S\ref{appendix:results}, Tab.~\ref{tab:baselines}, showing that these are strong baselines and \compressionrate{} is correlated with length). However, for prompted models, longer sequences are more challenging and \pass{} decreases with length. Overall, these results suggest that while scaling to large models is beneficial, generalization to long sequences and from synthetic to real distributions remains a major challenge.

\subsection{Additional Analyses}
\label{subsec:analysis}

\begin{table}[t]
  \begin{minipage}[t]{0.45\linewidth}
    \centering
    \footnotesize
    \begin{tabular}{l|c|cc}
       & \gzip{} & \multicolumn{2}{c}{\textsc{Llama-405B}} \\
      Length & \textsf{CR} & $\textsf{Acc.}$ &  $\textsf{Prec.}$ \\ \toprule 
      16 & 0.966 & \textbf{47.1} & 3.20 \\
      32 & 0.846 & 29.9 & 2.70 \\
      64 & 0.736 & 20.9 & 1.82 \\
      128 & 0.622 & 14.2 & \textbf{1.67} \\ \midrule
      1M & \textbf{0.398} & n/a & n/a \\ \bottomrule
    \end{tabular}
    \caption{\textbf{Effect of input length on Audio-8-bit sequences.} Performance improves for \gzip{} and \compressionrate{} as lengths increase, but \pass{} decreases.}
    \label{tab:length_audio}
  \end{minipage}
  \hfill
  \begin{minipage}[t]{0.45\linewidth}
    \centering
    \footnotesize
    \begin{tabular}{l|c|c}
      & Syn. & Audio  \\
      \toprule 
      \textsc{SQ-1.5B} & \textbf{86.6}  & 6.5 \\
      $+$ feed only &  77.6 & 8.6 \\
      $+$ feed \& ex &  75.9 & \textbf{8.7} \\ \bottomrule
    \end{tabular}
    \caption{\textbf{\pass{} for synthetic sequences and Audio MFCC sequences of length 64 bytes with and without execution feedback.} Execution feedback is not sufficient to close the gap between synthetic and real data.}
    \label{tab:ex_feedback}
  \end{minipage}
\end{table}

\paragraph{Effect of execution feedback.}
To examine the effect of adding execution feedback \citep{yang2023intercode, shojaee2023executionbasedcodegenerationusing, ni2024nextteachinglargelanguage}, we experiment with two variants: (a) \textbf{feedback only during training}: execution feedback is added to training examples, and (b) \textbf{feedback \& execution}: the code is executed by an interpreter at each line and results are presented to the model before the next step. In both cases the feedback is added as an end-of-line comment (see \S\ref{appendix:analysis}, Fig.~\ref{fig:example-ex-feedback}).

Tab.~\ref{tab:ex_feedback} presents our results for 1,280 synthetic and 1,280 Audio-MFCC sequences of length 64.  Execution feedback leads to small gains on real data, but also decreases performance on synthetic data and is not sufficient to significantly improve performance. This can be partially attributed to the fact that most training is spent on feedback tokens and the model is only trained on gold programs, and hence is not trained to dynamically fix model errors or bad trajectories \citep{ni2024nextteachinglargelanguage}.

\paragraph{Repetitions of input data.}
Python is highly expressive -- there are many ways the model can repeat input sequences (e.g., by extracting to multiple sub-sequences that are later concatenated, see \S\ref{appendix:analysis}, Fig.~\ref{fig:ex-repeat-success} and Fig.~\ref{fig:ex-err-repeat}). To check how often repetitions happen, we perform a qualitative analysis for 25 correct programs per modality (chosen randomly) for our \llamaxl{} and \gpt{} prompted models, presented in Tab.~\ref{analysis:repetitions}. We find that repetitions account for most correct predictions on natural data (more than $95\%$ on average), but only $20\%$ on our synthetic sequences, where models are able to leverage the synthetically induced patterns (Fig.~\ref{fig:ex-synthetic-success}). Hence, the low \precision{} of our prompted models can be partly attributed to the \gzip{} prior being under-optimized to Python programs.

\paragraph{Error analysis.}
To understand the errors our models make, we manually analyze 25 errors per modality for \llamaxl{} and \gpt{}, presented in Tab.~\ref{analysis:errors}. For both natural and synthetic sequences, most errors are caused by models misusing specific operators during execution, e.g., the wrong number of repetitions for a character (see Fig.~\ref{fig:ex-execution-err} and Fig.~\ref{fig:ex-err-syn} in \S\ref{appendix:analysis}). This can be attributed to the fact that current LLMs still struggle with symbolic reasoning and counting in particular \citep{hendrycksmath2021, yehudai2024transformerscountn, ball2024can}. 

Additionally, we identify a couple of specific failure cases. In more than 30\% of cases, models try repeating the input sequence by breaking it to sub-sequences that are later merged, but still introduce subtle errors when constructing and merging these sub-sequences (Fig.~\ref{fig:ex-repeating-err}).
We identify a special failure case for the text modality, where models successfully reverse-engineer our data generation process and generate a string which they convert to UTF-8-encoded bytes, but often with subtle errors leading to low success rates (Fig.~\ref{fig:ex-text-err}).
\section{Related Work}

\paragraph{Data compression.}
Data compression is one of the main problems in computer science \& communications. In addition to the works described in \S\ref{sec:background}, fundamental work includes \citet{6773024, 10.5555/1146355}. See \citet{salomon2004data, 10.1145/45072.45074, JAYASANKAR2021119} for detailed surveys. A major line of work is devoted towards domain-specific algorithms, including specifically for text, audio, and DNA \citep{10.1145/322344.322346, 10.1145/1753171.1753177,salau2019audio, du2020compression}. Recently, it has been proposed that LLMs can be harnessed for compression via arithmetic coding \citep{valmeekam2023llmziplosslesstextcompression, deletang2024language}. While the raw compression rate (excluding the weights of the LLM) of this method outperforms classical compression algorithms, the full LLM parameters are required for decoding. Our work differs by prompting and training \codegenllms{} to directly generate \emph{programs} that produce the original data.

\paragraph{Evaluation of \codegenllms{}.}
Recent jumps in the coding ability of LLMs \citep{chen2021evaluatinglargelanguagemodels, rozière2024codellamaopenfoundation} paved the path to benchmarks that focus on code generation. For example, \citet{chen2021evaluatinglargelanguagemodels} on generating code given a function's signature, \citet{austin2021programsynthesislargelanguage} on solving coding interview questions, and \citet{doi:10.1126/science.abq1158} focus on competitive programming tasks. Other benchmarks focus on agentic settings where models interact with an environment to solve GitHub issues or run research experiments \citep{jimenez2024swebenchlanguagemodelsresolve, bogin2024superevaluatingagentssetting}. KT has several advantages over previous benchmarks: there is an abundance of diverse natural data, strong baselines for compression already exist, and evaluation metrics are well-defined. Our work suggests that making progress on KT is challenging and new innovations are needed for further progress.
Moreover, to outperform current compression methods, \codegenllms{} must model complex environments, potentially improving their ability to act as world models \citep{tang2024worldcoder}.

\paragraph{Synthetic data augmentation for coding and symbolic reasoning.}
Synthetic data augmentation has been recently shown to be highly effective in injecting numerical skills to LMs \citep{geva2020injectingnumericalreasoningskills, yoran-etal-2022-turning, lu2024mathgeniegeneratingsyntheticdata, mitra2024orcamathunlockingpotentialslms}, however improvements do not necessarily generalize \citep{liu2023tinygsmachieving80gsm8k}. More complex pipelines for synthetic data generation with LLMs are often used for \codegenllms{} \citep{rozière2024codellamaopenfoundation}.
\section{Discussion and Limitations}
\label{discussion}

\paragraph{Generalization from synthetic to real data.}
Our results show that generalization from synthetic to real distributions remains a challenge. Future work can experiment with methods that tilt the synthetic distribution towards the real one, e.g., by filtering easy-to-detect synthetic sequences \citep{ganin2015unsuperviseddomainadaptationbackpropagation}. An exciting direction for future research is to model KT as a reinforcement learning task \citep{Fran_ois_Lavet_2018, mnih2013playingatarideepreinforcement} and learn from a reward when correct solutions are generated, with a bias towards shorter solutions \citep{ellis2020dreamcodergrowinggeneralizableinterpretable}. Additionally, future methods can benefit from curriculum-learning where models learn simpler examples before more complex ones \citep{hacohen2019powercurriculumlearningtraining, 10.1007/s11263-022-01611-x}.

\paragraph{Additional modalities.}
This work focuses on natural text, audio, and DNA sequences in addition to our synthetic data, and can be easily extended to cover additional modalities such as publicly-available image and video sequences \citep{5206848, soomro2012ucf101dataset101human}. An exciting direction for future work is to mix training examples from different domains, composing of different patterns and functionalities. Additionally, we only focus on the ability of \codegenllms{} to compress sequences of data, and leave compression of executable programs for future work.

\paragraph{Compressing larger amounts of data.} In this work, we focus on compressing 1MB of data, a smaller amount than previous work \citep{deletang2024language} and the Hutter Prize \citep{hutter2009hutter}, which focus on 1GB of data per modality. We opt for 1MB sequences as this amounts to roughly 60K sequences across all modalities (around 20K for 1MB of DNA and 7800 sequences for each of the other modalities), a relatively large amount of examples in terms of current code generation benchmarks \citep{chen2021evaluatinglargelanguagemodels, austin2021programsynthesislargelanguage, doi:10.1126/science.abq1158, li2023tacotopicsalgorithmiccode}. Nevertheless, compressing larger sequences is likely to be feasible as \codegenllms{} and hardware continue to improve. To assist further progress, we will also release 1GB versions of our datasets.

\paragraph{Scaling to longer sequences.}
In our synthetic experiments, we set the maximum size of a sub-sequence to 25, and the maximum number of initiated sub-sequences to 5 (see \S\ref{appendix:syn_data} for all hyper-parameters). Hence, a program will not produce a sequence of more than 125 random characters (concatenating 5 sub-sequences of 25 random characters), but can produce longer sequences by using \emph{modifiers}, e.g., by repeating a sub-sequence. Extending our DSL to longer sequences, for example by further scaling training compute, can be an interesting direction for future research. 

Moreover, longer sequences sequences yield lower compression rates, as exemplified for \gzip{} in Tab.~\ref{tab:length_audio}. Although we mainly experiment with sequences of length 128 due to the low \pass{} of current models on long sequences, future models might be able to achieve better compression on much longer sequences.

\paragraph{Other languages}
We have experimented with general Python as well as a simple sequence-oriented DSL.
Future work could explore how the choice of language affects the performance of models, the rate of training progress, and the degree of generalization to OOD sequences.

\paragraph{Stronger priors.}
In our experiments we used the \gzip{} prior for programs from our prompted models and our uniform prior over our synthetic DSL. Future work can experiment with stronger priors, e.g., small LMs that are trained on the distribution of programs. Another possible future direction is to focus on lower level languages where developing strong priors might be simpler. Although such programs might be harder for humans to understand, the potential of \codegenllms{} to generate novel solutions is an exciting opportunity for future work.

\paragraph{Runtime compute.}
In our experiments, we do not consider the runtime compute of programs execution. To limit the effect of inefficient programs, we limit the runtime of each program to five seconds and consider programs that run for longer as \emph{non-executable} (similar to runtime exceptions). We provide additional statistics on failed and non-executable programs in \S\ref{appendix:results}. Similarly, we do not consider the weights of the LLM (which are only required for generating the programs) or the cost of the Python standard library required to execute the programs. Additionally, we leave experimenting with specialized tokenization methods, e.g., \citet{dagan2024gettingtokenizerpretrainingdomain}, for future work.

\section{Conclusion}
In this work we introduce the \kt{}, a compression-as-intelligence test for code generation models, based on the notion of \emph{Kolmogorov complexity}. To evaluate current models, we use real audio, text, and DNA data, and develop an automatic framework for sampling program-sequence pairs. KT is extremely challenging for current models, including \llamaxl{} and \gpt{}. Additionally, for synthetic distributions where pairs can be automatically sampled, trained models have a better \compressionrate{} than current methods. As our method can potentially be scaled to an infinite number of real problems of varying difficulty, it can provide  a challenging evaluation test-bed for future code generation models and pave the path towards new methods that compress large amounts of real data.
\section*{Reproducibility}
We have made significant efforts to ensure our work can be reproduced. Our datasets will be fully released, including our DSL and synthetic data generation framework. We will also release larger 1GB variants to facilitate future research. In addition, we will release code to reproduce results with our prompted models and will host a live leaderboard to allow the community to monitor future developments. As we will not be able to open-weight our \seqcoder{} models, we will release training code for \seqcodersmall{} models on our synthetic pairs, enabling reproduction of our results and simplify training new \seqcoder{} models.

\section*{Ethics Statement}
In this work, we introduce the \kt{}, a challenging intelligence test for \codegenllms{} based on the notion of Kolmogorov complexity. While we do not see any immediate risk caused by KT, code generation research is not without risk. Powerful models can be used for malicious reasons, e.g., by hackers or other malicious users.
If compression via zero-shot code generation was solved, it might enable a leap in telecommunications efficiency while at the same time raising concerns about opaque behavior of strong artificial intelligence.

\bibliography{iclr2025_conference}
\bibliographystyle{iclr2025_conference}

\appendix
\section{Data Generation}
\label{appendix:data}

\paragraph{Additional details and statistics.}
We randomly choose audio sequences from LibriSpeech until reaching 1MB of data. As we opt not to trim sequences, we are left with 1,015,360 bytes in the 16-bit setting, 1,00,3280 bytes in the 8-bit setting, and 1,000,960 bytes in the MFCC setting. Similarly, we have 1,004,232 text bytes, and 1,000,105 synthetic bytes. For DNA, we have exactly 1MB.

The maximum sequence length shown to our models is 128. As we do not merge between different origin sequences (i.e., an audio file or synthetic sequence), some sequences can be shorter. The average sequence length for natural sequences is 75.9 for our synthetic data, 126.0 for MFCC and 127.9 for all other modalities. We provide detailed statistics regarding synthetic program-sequence lenghts in \S\ref{appendix:syn_data}.

\paragraph{Figures.}
The DNA illustration in Fig.~\ref{fig:fig2} is courtesy of The National Human Genome Research Institute\footnote{\url{https://www.genome.gov/genetics-glossary/acgt}}.

\subsection{Naturally Occurring Sequences}
\label{appendix:natural}

\paragraph{Audio MFCC.}
We use Librosa\footnote{\url{https://librosa.org/doc/main/generated/librosa.feature.mfcc.html}} to parse our audio files to MFCC. Specifically, we parse each file to 20 MFCCs. As MFCCs are floats, we round each float into an integer which we take modulo 256 to obtain numbers in the range $[0, 255]$. Note that this representation is far from being reconstructible but we posit that it provides a source of richly structured natural data nevertheless.

\subsection{Synthetic Data Generation}
\label{appendix:syn_data}

\paragraph{Automatic generation of program-sequence pairs.}
Our DSL supports the following functions and can be easily extended in the future:
\begin{itemize}
    \item \textbf{[Initiator] \emph{Set list}}: Initiates a fixed sequence of numbers.
    \item \textbf{[Initiator] \emph{Range}}: Initiates a sequence of increasing numbers from start to end index, either with or without a step interval.
    \item \textbf{[Initiator] \emph{Repeat}}: Initiates a sequence of repeated occurrences of the same number.
    \item \textbf{[Modifier] \emph{Substitute}}: Substitutes a value in a list with a different value.
    \item \textbf{[Modifier] \emph{Reverse}}: Reverses a sequence.
    \item \textbf{[Modifier] \emph{Sub-sequence between indexes}}: Returns a sub-sequence between two indexes, either with or without intervals.
    \item \textbf{[Modifier] \emph{Repeat list}}: Returns a repetition of a sequence a specified number of times.
    \item \textbf{[Modifier] \emph{Max/Min}}: Returns the $n$ maximum or minimum items in the sequence.
    \item \textbf{[Modifier] \emph{Add/Subtract/Modulo}}: Applies a mathematical operation (addition, subtraction, or modulo) with a specified fixed operand to all elements in the sequence.
    \item \textbf{[Modifier] \emph{Scan add}}: Returns a new sequence where the value at index $n$ is the sum of all items up to that index.
    \item \textbf{[Filter] \emph{Is even/odd}}: Keeps only even or odd numbers in the sequence.
    \item \textbf{[Filter] \emph{Is non-zero}}: Keeps only non-zero numbers in the sequence.
    \item \textbf{[Merger] \emph{Add/Subtract/Modulo two lists}}: Applies a mathematical operation (addition, subtraction, or modulo) over elements at matching indexes in two sequences.
    \item \textbf{[Merger] \emph{Concatenate}}: Concatenates two sequences.
    \item \textbf{[Merger] \emph{Interleave}}: Interleaves one sequence with another in alternating indexes. If one sequence is longer, the remaining values will form the suffix of the interleaved sequence.
\end{itemize}

When sampling from our DSL, we first sample the initiators, then produce additional sequences using sampled filters and modifiers before the sub-sequences are composed by sampled mergers to form the final sequence. We can control the distribution of programs via the hyper-parameters of the sampling process.

The number of initiators is uniformly sampled in range $\{1, ..., 5\}$, the length of each fixed sequence is sampled uniformly in range $\{5, ..., 25\}$. 

The probability to apply any non-mathematical modifier (e.g., \emph{repeat} or \emph{substitute}) is set to $0.4$, and the probability for each specific modifier is set to $0.1$. After the modification, we reuse the original sequence at probability $0.2$. For substitution, we exclude the original sequence at probability $0.25$. Similarly, the probability to apply any mathematical modifier (e.g., \emph{Add/Subtract/Modulo}) to an initiated sequence or filter is set to $0.4$, and the probability for a specific filter is set to $0.1$. Mathematical mergers are treated similarly -- the probability to any mathematical merger is $0.4$ and for a specific merger is $0.1$. Finally, the probability for concatenate and interleave operations is set to $0.8$ and $0.2$, respectively, and we continue to concatenate and interleave sub-sequence until all sub-sequences are used in the final output sequence. 

Additionally, we make sure that sequences in our evaluation data will not be part of a program-sequence pair during training. In case a sequence is generated by more than one program, we only keep the shortest program that produces the sequence in our dataset.

\begin{figure}[ht]
  \centering
  \begin{minipage}{0.45\textwidth}
    \centering
    \includegraphics[width=\textwidth]{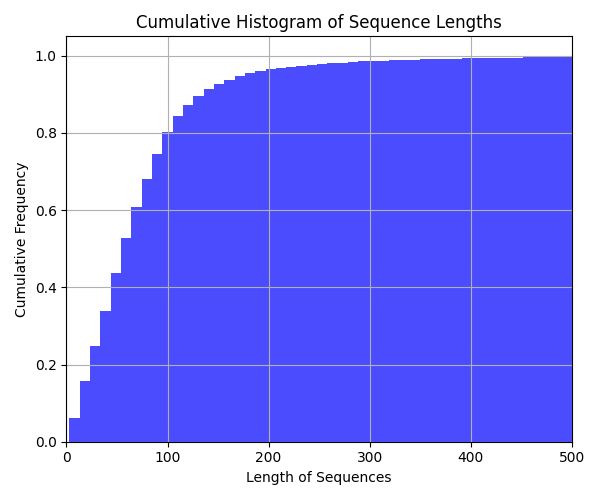}
    \caption{Cumulative histogram of sequence lengths for our synthetic distribution.}
    \label{fig:syn_seq_length_hist}
  \end{minipage}%
  \hfill
  \begin{minipage}{0.45\textwidth}
    \centering
    \includegraphics[width=\textwidth]{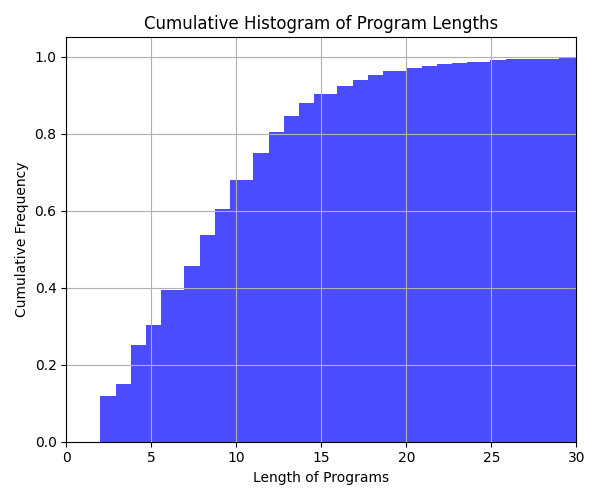}
    \caption{Cumulative histogram of program lengths (number of lines) for our synthetic distribution.}
    \label{fig:program_length}
  \end{minipage}
\end{figure}

\paragraph{Program-sequence lengths.}
Fig.~\ref{fig:syn_seq_length_hist} and Fig.~\ref{fig:program_length} presents the histogram of the lengths of our synthetic sequences and programs respectively.

\paragraph{Uniform prior over DSL programs.}
Alg.~\ref{alg:uni-encoding} presents the number of bits needed to sample a program using the uniform prior over programs and arithmetic coding \citep{5391119, 1055739}.
The cost to encode each line has two main components. 

The first, is to encode the method that is being called. Since we use a uniform prior, this is always $log$ of the number of functions (stopping the generation and assigning one of the previously defined sequences as the output is formulized as an additional function). 

The second, is to encode the variables that are being passed to the function, which is determined by the environment and is specific for each function. For example, setting a random list of elements (\emph{set list}), requires $l*b$ bytes, $l$ indicating the length of the list, and $b$ indicating the number of bits in each byte. Similarly, concatenation two previously defined sub-sequences requires encoding the indexes of these sequences (which is a $log$ operation of the current number of lines), and repeating a character requires encoding the character (again, the byte size) and the number of repetitions ($log$ of the maximal number of allowed repetitions in the environment).

\begin{algorithm}[H]
\footnotesize 
\begin{algorithmic}
\Function{encode\_func\_params}{func, func\_params, hps}
    \State \textbf{mapping} = \{ 
        “set\_list”: $len(func\_params.list\_size) \times hps.byte\_size$, 
        “concatenate”: $\log_2(hps.line\_index) \times 2$, 
        “repeat\_num$: hps.byte\_size + hps.max\_num\_repetitions$, ...
    \}
    \State \Return mapping[func]
\EndFunction

\State $x \leftarrow 0$
\State $bits\_per\_func \leftarrow \log_2(num\_functions+1)$

\ForAll{$i, (f, f\_params)$ in enumerate(program\_lines)} 
    \State $x \mathrel{+}= bits\_per\_func$ \Comment{\# bits to encode the function type}
    \State $x \mathrel{+}= encode\_func\_params(f, f\_params, hps, i)$ \Comment{\# bits to encode the function params}
\EndFor
\State \Return $x$
\end{algorithmic}
\caption{Encoding a program under the uniform prior.}
\label{alg:uni-encoding}
\end{algorithm}

\subsection{Formal Definition of the \kt{}}
\label{subsec:kt_alg}

Alg.~\ref{alg:kt} presents a formal definition of the \kt{}. \codegenllms{} that generate shorter programs have better compression rates (see \S\ref{sec:exps} for our metrics and experimental settings).

\begin{algorithm}[H]
\caption{Compression by Code Generation (The \kt{})}
\begin{algorithmic}[1]
\Require Sequence $x$, \codegenllm{} $M$, Programs prior $p$

\State Generate program $\rho$ with model $M$ that produces the sequence $x$ 
\State Encode $\rho$ with prior $p$
\State \Return $p(\rho)$
\end{algorithmic}
\label{alg:kt}
\end{algorithm}

\section{Experiments}

\subsection{Models}
\label{appendix:models}

\paragraph{\gzip{} baseline.}
We use the python implementation of \gzip{}\footnote{\url{https://docs.python.org/3/library/gzip.html}}. To enable reconstruction of the sequential data, we use new line as the separator between sequences.

\paragraph{\lmis{} baseline.}
We use our \seqcodertiny{} model for the results reported in \S\ref{sec:results} and provide additional results in \S\ref{appendix:results}.

\begin{figure}[ht]
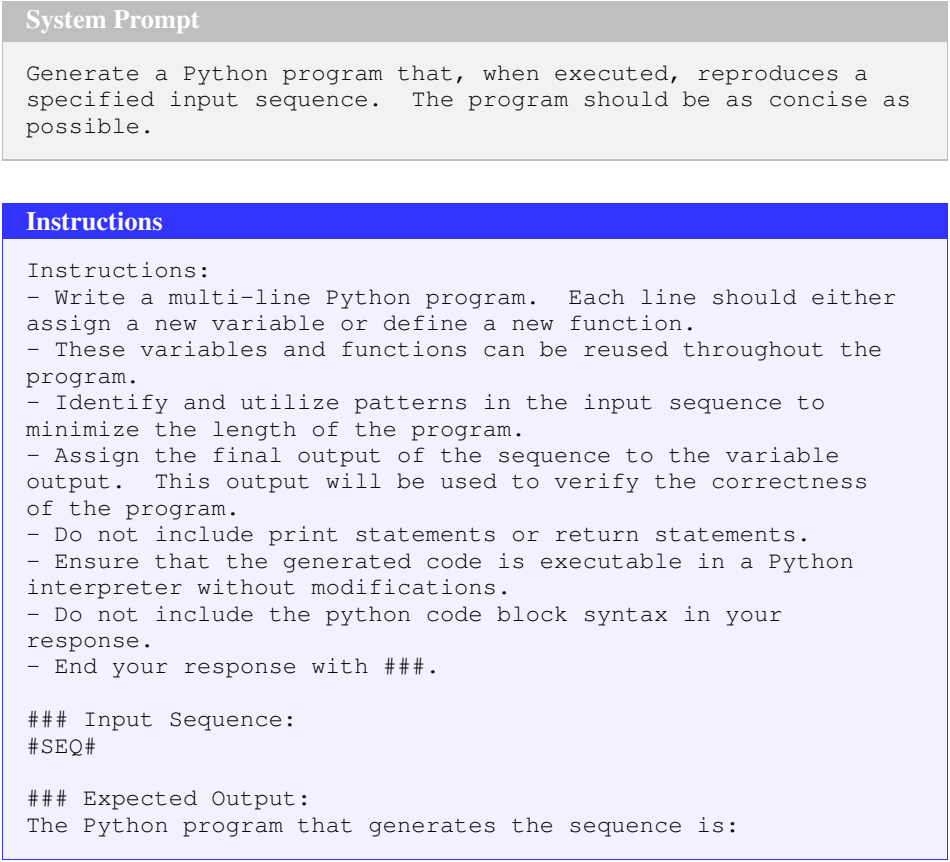

    \centering
    \begin{tcolorbox}[
        colback=gray!10,
        colframe=gray!50,
        title=System Prompt,
        width=0.9\textwidth,
        boxrule=0.5pt, 
        left=2mm, 
        right=2mm,
        top=2mm,
        bottom=2mm,
        fonttitle=\bfseries,
        sharp corners, 
        ]
        \ttfamily \small
        Generate a Python program that, when executed, reproduces a specified input sequence. 
        The program should be as concise as possible.
    \end{tcolorbox}
    
    \vspace{5pt}
    \begin{tcolorbox}[
        colback=blue!5,
        colframe=blue!80,
        title=Instructions,
        width=0.9\textwidth,
        boxrule=0.5pt, 
        left=2mm, 
        right=2mm,
        top=2mm,
        bottom=2mm,
        fonttitle=\bfseries,
        sharp corners
        ]
        \ttfamily \small 
        Instructions:\newline
        - Write a multi-line Python program. Each line should either assign a new variable or define a new function. \newline
        - These variables and functions can be reused throughout the program.\newline
        - Identify and utilize patterns in the input sequence to minimize the length of the program.\newline
        - Assign the final output of the sequence to the variable \texttt{output}. This output will be used to verify the correctness of the program. \newline
        - Do not include print statements or return statements.\newline
        - Ensure that the generated code is executable in a Python interpreter without modifications.\newline
        - Do not include the \texttt{python} code block syntax in your response.\newline
        - End your response with \texttt{\#\#\#}.\newline
        
        \#\#\# Input Sequence:\newline
        \texttt{\#SEQ\#} \newline
        
        \#\#\# Expected Output:\newline
        The Python program that generates the sequence is:
    \end{tcolorbox}

    \caption{Full prompt used in our zero-shot experiments.}
    \label{fig:prompt}
\end{figure}

\paragraph{Prompted baselines.}
Fig.~\ref{fig:prompt} presents our full prompt used in our prompted models experiments. We use vLLM\footnote{https://github.com/vllm-project/vllm} to set-up inference servers with the instruction-tuned \textsc{Llama-3.1} models available on HuggingFace\footnote{https://huggingface.co/meta-llama/Meta-Llama-3.1-\{8B/70B/405B\}}.

\begin{figure}[ht]
    \centering
    \begin{tcolorbox}[
        colback=gray!10,
        colframe=gray!50,
        title=System Prompt,
        width=0.9\textwidth,
        boxrule=0.5pt, 
        left=2mm,
        right=2mm,
        top=2mm,
        bottom=2mm,
        fonttitle=\bfseries,
        sharp corners,
        ]
        \ttfamily \small 
        Generate a Python program that, when executed, reproduces a specified input sequence. 
        The program should be as concise as possible.
    \end{tcolorbox}
    
    \vspace{5pt}
    
    \begin{tcolorbox}[
        colback=blue!5,
        colframe=blue!80,
        title=Instructions,
        width=0.9\textwidth,
        boxrule=0.5pt,
        left=2mm, 
        right=2mm,
        top=2mm,
        bottom=2mm,
        fonttitle=\bfseries,
        sharp corners
        ]
        \ttfamily \small
        Instructions:\newline
        - Write a multi-line Python program. Each line should either assign a new variable or define a new function. \newline
        - These variables and functions can be reused throughout the program.\newline
        - Identify and utilize patterns in the input sequence to minimize the length of the program.\newline
        - Assign the final output of the sequence to the variable \texttt{output}. This output will be used to verify the correctness of the program. \newline
        - Do not include print statements or return statements.\newline
        - Ensure that the generated code is executable in a Python interpreter without modifications.\newline
        - Do not include the \texttt{python} code block syntax in your response.\newline
        - End your response with \texttt{\#\#\#}.\newline
        - \textbf{Before the program, you can use the Thought field to generate how you think the task should be solved. After the thought, generate "The Python program that generates the sequence is:", followed by the program.}
        \newline
        \newline
        \#\#\# Input Sequence:\newline
        \texttt{\#SEQ\#} \newline
        
        \#\#\# Expected Output:\newline
        The Python program that generates the sequence is:
    \end{tcolorbox}

    \caption{Full prompt used in our zero-shot CoT experiments. The difference from our non-CoT prompt in Fig.~\ref{fig:prompt} is highlighted.}
    \label{fig:prompt-cot}
\end{figure}

\paragraph{Trained models.}
Our \seqcodertiny{} model is a transformer \citep{10.5555/3295222.3295349} with 1.5 billion parameters. The model follows a similar architecture to the recent \textsc{Llama} models \citep{touvron2023llama2openfoundation, dubey2024llama3herdmodels} but with less parameters -- it has 24 layers, a hidden-dimension of 2,048, multi-head attention with 16 heads, and a maximum sequence length of 2,048 tokens. To endow the model with coding and reasoning skills before training on synthetic data from our DSL, we pre-train on 200B tokens of publicly-available text and code. Although this model is under-trained in relation to current state-of-the-art models and hence significantly weaker, it achieves non-negligible performance on popular coding and reasoning tasks -- \textsf{Pass@1} of $11.6$ on HumanEvalPlus \citep{evalplus}, and \pass{} of $28.2\%$ on ARC \citep{allenai:arc} and $68.2\%$ on PIQA \citep{bisk2019piqareasoningphysicalcommonsense}.

\newpage
\subsection{Results}
\label{appendix:results}

\begin{table*}[t]
\centering
\tiny
\begin{tabular}{l|cc|cc|cc|cc|cc|cc}
\toprule
& \multicolumn{2}{c}{Audio-16-bit} & \multicolumn{2}{c}{Audio-8-bit} & \multicolumn{2}{c}{Audio-MFCC} & \multicolumn{2}{c}{DNA} & \multicolumn{2}{c}{Text}  & \multicolumn{2}{c}{Syn.} \\ 
& $\textsf{Acc.}$ & $\textsf{Prec.}$ &  $\textsf{Acc.}$ & $\textsf{Prec.}$ &  $\textsf{Acc.}$ & $\textsf{Prec.}$ &  $\textsf{Acc.}$ & $\textsf{Prec.}$ &  $\textsf{Acc.}$ & $\textsf{Prec.}$ &  $\textsf{Acc.}$ & $\textsf{Prec.}$\\ 
\midrule
\llamasmall& 2.9 & 1.6 & 1.60 & 2.05 & 5.4 & 1.59 & 6.9 & 3.8 & 0.4 & 2.67 &  6.9 & 3.8 \\
\llamalarge & 14.3 & 1.61 & 4.2 & 1.74 & 23.6 & 1.50 & 9.2 & 2.25 & 3.7 & 2.69 & 16.9 & 2.93 \\
\llamaxl & \textbf{37.9} & \textbf{1.53} & \textbf{10.4} & \textbf{1.60} & \textbf{49.7} & \textbf{1.41} & \textbf{23.4} & \textbf{1.94} & \textbf{5.1} & \textbf{2.35} & \textbf{30.5} & \textbf{2.6} \\
\bottomrule
\end{tabular}
\caption{\pass{} and \precision{} for zero-shot prompted models on the \kt{} for sequences of length $128$ with CoT. Trends are similar to those without CoT in Tab.~\ref{tab:results_zero_shot}.}
\label{tab:results_zero_shot_cot}
\end{table*}

\begin{table*}[t]
\centering
\footnotesize
\begin{tabular}{l|ccc|ccc}
\toprule
& \multicolumn{3}{c}{\gpt{}} & \multicolumn{3}{c}{\omini{}} \\ 
& \textsf{\% Ex.} & \textsf{Ex. Acc.} & \textsf{Acc.} & \textsf{\% Ex.} & \textsf{Ex. Acc.} & \textsf{Acc.} \\ 
\midrule
Audio-16-bit & 88.2 & 78.8 & 69.5 & 81.2 & 87.7 & 71.2 \\
Audio-8-bit & 94.6 & 38.5 & 36.4 & 82.2 & 51.1 & 42.0 \\
Audio-MFCC & 94.4 & 88.8 & 83.8 & 86.8 & 93.8 & 81.4 \\
DNA & 93.1 & 54.0 & 50.3 & 85.6 & 67.1 & 57.4 \\
Text & 76.5 & 70.8 & 54.2 & 82.8 & 53.6 &  44.4\\
Synthetic & 92.9 & 48.1 & 44.7 & 71.4 & 65.8 & 47.0 \\ \midrule
Average & 89.9 & 63.2 & 56.5 & 81.7 & 69.8 & 57.2 \\
\bottomrule
\end{tabular}
\caption{The percentage of executable programs (\textsf{\% Ex.}), accuracy for executable programs (\textsf{Ex. Acc.}), and \pass{} (\textsf{Acc.}) for \gpt{} and \omini{}. To reduce costs, \omini{} was evaluated on 500 random sub-sequences of length 128. On average, both models have similar \pass{}. \omini{} has higher accuracy for executable programs, but a lower percentage of executable programs.}
\label{tab:gpts}
\end{table*}

\begin{table}[ht]
\footnotesize
\centering
\begin{tabular}{l|cc|ccc}
\toprule
 & \multicolumn{2}{c|}{\lmis{}} & \multicolumn{3}{c}{\gzip{}} \\ 
Sequence length & 128 & 1024 & 128 & 1024 & 1MB \\ \midrule
Audio-16-bit & 0.697 & \textbf{0.639} & 1.147 & 1.014 & 0.920 \\
Audio-8-bits & 0.319 & \textbf{0.273} & 0.621 & 0.454 & 0.398 \\
Audio MFCC & 0.739 & \textbf{0.698} & 1.250 & 0.974 & 0.903 \\
DNA & 0.750 & \textbf{0.662} & 1.182 & 0.894 & 0.714 \\ 
Text & 0.611 & 0.463 & 0.782 & 0.525 & \textbf{0.357} \\ 
Synthetic & \textbf{0.449} & 0.450 & 0.943 & 0.819 & 0.593 \\ \midrule
Average & 0.594 & \textbf{0.531} & 0.988 & 0.748 & 0.647 \\ \bottomrule
\end{tabular}
\caption{\compressionrate{} for \gzip{} and \lmis{} with various input sequence lengths.}
\label{tab:baselines}
\end{table}

\paragraph{Chain-of-thought prompting does not lead to additional improvements.} We present our CoT prompt in Fig.~\ref{fig:prompt-cot} and our experiments with CoT with \textsc{Llama} in Tab.\ref{tab:results_zero_shot_cot}.
Similarly to recent work \cite{sprague2024cotcotchainofthoughthelps, zheng2024makeslargelanguagemodels}, we do not observe significant gains with CoT prompting.

We present the percentage of executable programs, accuracy for executable programs, and \pass{} for \gpt{} and \omini{} in Tab.~\ref{tab:gpts}.
While \omini{} has a higher accuracy for executable programs (69.8 vs 63.2), it also has a lower percentage of executable programs (81.7 vs 89.9), and both models have similar \pass{} (57.2 vs 56.5), suggesting that while longer chains-of-thoughts are able to increase performance, they do not yet lead to major improvements on the \kt{}.

\paragraph{\gzip{} and \lmis{} are strong baselines.}
Tab.~\ref{tab:baselines} presents results for our \gzip{} and \lmis{} baselines, for sequences of length 128 and 1,024\footnote{Each element in the sequence is a single token, and we use a comma separator between elements. We do not consider the cost of generating the separator token for the \lmis{} baseline.}. For \gzip{} we also compress all 1MB as a single sequence. We note that stronger compression baselines exist, and that the current state-of-the-art \compressionrate{} for our text data from the Hutter Prize is around 0.11 \citep{hutter2009hutter}.

\lmis{} is a strong baseline, it outperforms \gzip{} (with 1MB) on $5/6$ modalities with a sequence length of 1,024 and $4/6$ modalities with a sequence length of 128. Our results on Audio-8-bit are comparable to the \gzip{} and \textsc{Chinchilla-1B} baseline from \citet{deletang2024language} (which reports compression rates of 0.249 with \lmis{} and 0.364 for \gzip{} with 1GB of data, both within a
 a $3.6$-point margin).
On on text, our \lmis{} baseline is significantly weaker than the one in \citet{deletang2024language}. We attribute the difference to the different setting -- we present the sequences as \emph{decimal numbers} (to enable model to identify mathematical patterns), while the original work presents sequences as \emph{ASCII characters}. Moreover, \gzip{} improves as we increase the input length, reaching \compressionrate{} of 0.32 for 1GB of data, similar to \citet{deletang2024language}.

\paragraph{Scaling \seqcoder{} models improves \pass{} on synthetic data.}
Fig.~\ref{fig:pass_1_syn} presents the \pass{} for the different \seqcoder{} models on our held-out synthetic dataset. \seqcodertiny{} and \seqcodersmall{} models are trained for $20K$ and $10K$ steps, respectively. \seqcodertiny{-10K} overfits and accuracy decreases with additional training. \seqcodertiny{} (1M examples) significantly outperforms \seqcodertiny{-100K} and the larger \seqcodersmall{} performs better and learns faster. 

\paragraph{Scaling prompted models increases percentage of executable programs.}
Fig.~\ref{tab:ex_fails} shows the percentage of programs that execute without error and terminate within five seconds, broken down by model and dataset. Larger models generate more executable programs. Programs generated by \llamaxl{} and \gpt{} are executable $77.2\%$ and $89.9\%$  of the time on average across modalities.
 
\begin{table}[htbp]
  \centering
  \begin{minipage}{0.45\textwidth}
    \centering
    \includegraphics[width=\textwidth]{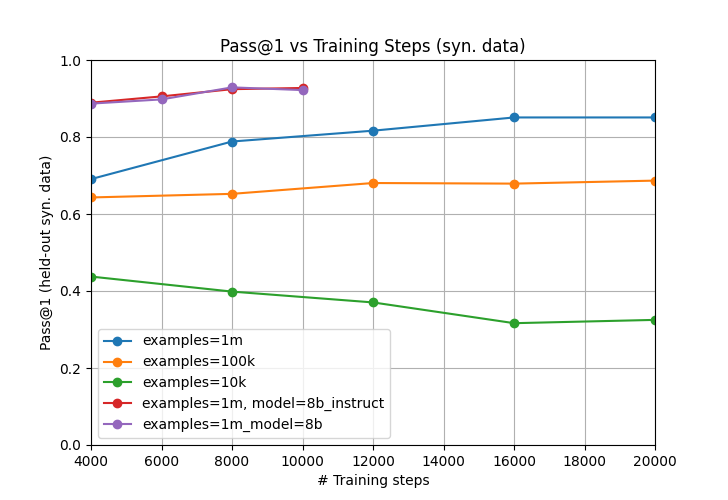}
    \caption{\pass{} on our synthetic held-out set for the different \seqcoder{} models.}
    \label{fig:pass_1_syn}
  \end{minipage}%
  \hfill
  \begin{minipage}{0.45\textwidth}
    \centering
    \footnotesize
    \begin{tabular}{l|ccc|c}
       & \multicolumn{3}{c}{\textsc{Llama-3.1}}  & \textsc{GPT}\\
      Modality & \textsc{8B} & $\textsc{70B}$ & $\textsc{405B}$ & \textsc{4-o} \\ \toprule 
      Audio-16-bit & 25.6 & 67.2 & 69.3 & \textbf{88.2}\\
      Audio-8-bit  & 28.2 & 82.0 & 87.4 & \textbf{94.6}\\
      Audio-MFCC  & 25.9 & 71.0 & 76.6 & \textbf{94.4}\\
      DNA & 33.7 & \textbf{93.9} & 91.9 & 93.1 \\ 
      Text & 4.3 & 51.0 & 49.1 & \textbf{76.5}\\ 
      Synthetic & 42.0 & 85.4 & 88.9 & \textbf{92.9}\\ \midrule
      Average & 26.6 & 75.1 & 77.2 & \textbf{89.9} \\ \bottomrule
    \end{tabular}
    \caption{The percentage of executable programs for our prompted models.}
    \label{tab:ex_fails}
  \end{minipage}
\end{table}
  
\paragraph{Strong priors are necessary for good compression.}
In \S\ref{sec:results}, Fig.~\ref{fig:sub1} we show that we can outperform previous compression methods for our synthetic data using \seqcodertiny{} and a uniform prior over programs and arithmetic coding. The same programs achieve a compression rate of $5.26$ and $0.59$ with raw (the size of the programs saved as strings on the disk) and \gzip{} encoding, showing that strong priors are needed for good \compressionrate{}.

\subsection{Analysis}
\label{appendix:analysis}

\begin{figure}[ht]
    \centering
    \begin{subfigure}[b]{0.45\textwidth}
        \centering
        \includegraphics[width=\textwidth]{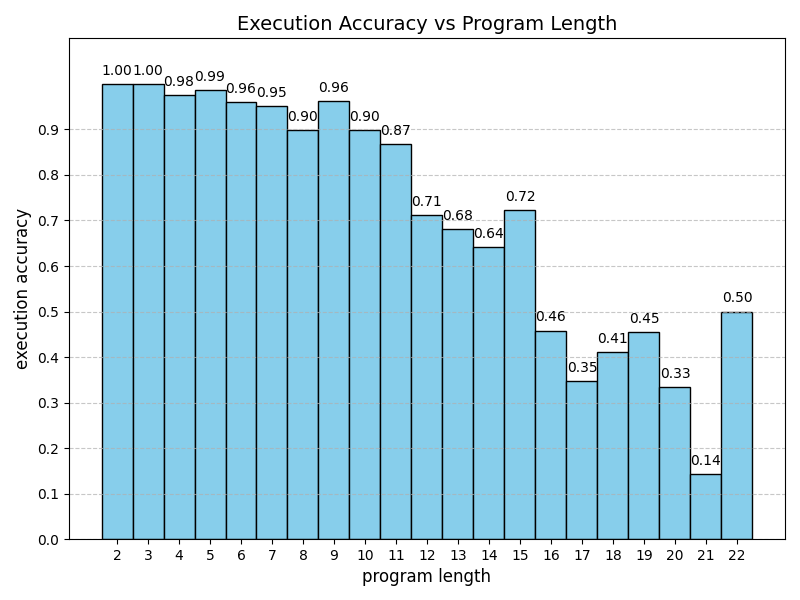}
        \caption{\seqcodertiny{} accuracy as a function of gold program lengths.}
        \label{fig:syn_program_lenghts}
    \end{subfigure}
    \hfill
    \begin{subfigure}[b]{0.45\textwidth}
        \centering
        \includegraphics[width=\textwidth]{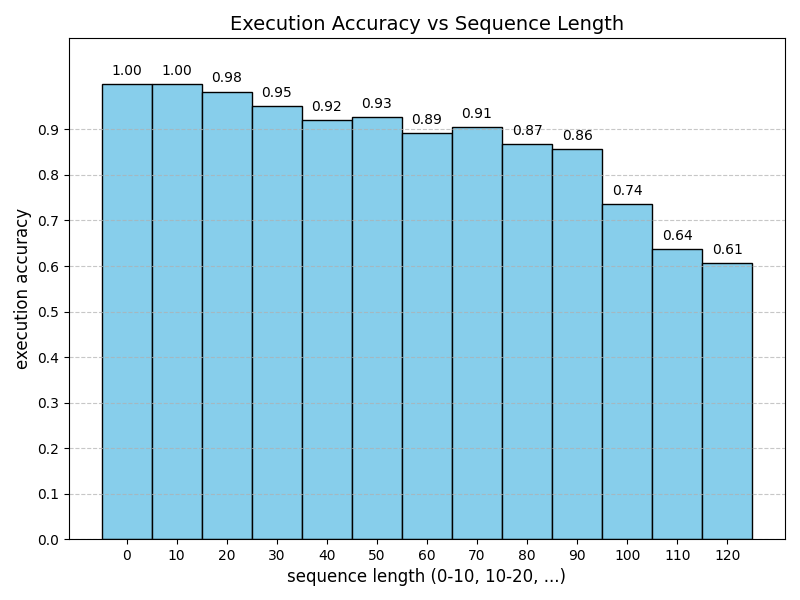}
        \caption{\seqcodertiny{} accuracy as a function of input sequence lengths.}
        \label{fig:syn_sequence_lenghts}
    \end{subfigure}
    \caption{\seqcodertiny{} accuracy as a function of gold program and input sequence lengths.}
    \label{fig:syn_lenghts}
\end{figure}

\begin{figure}[ht]
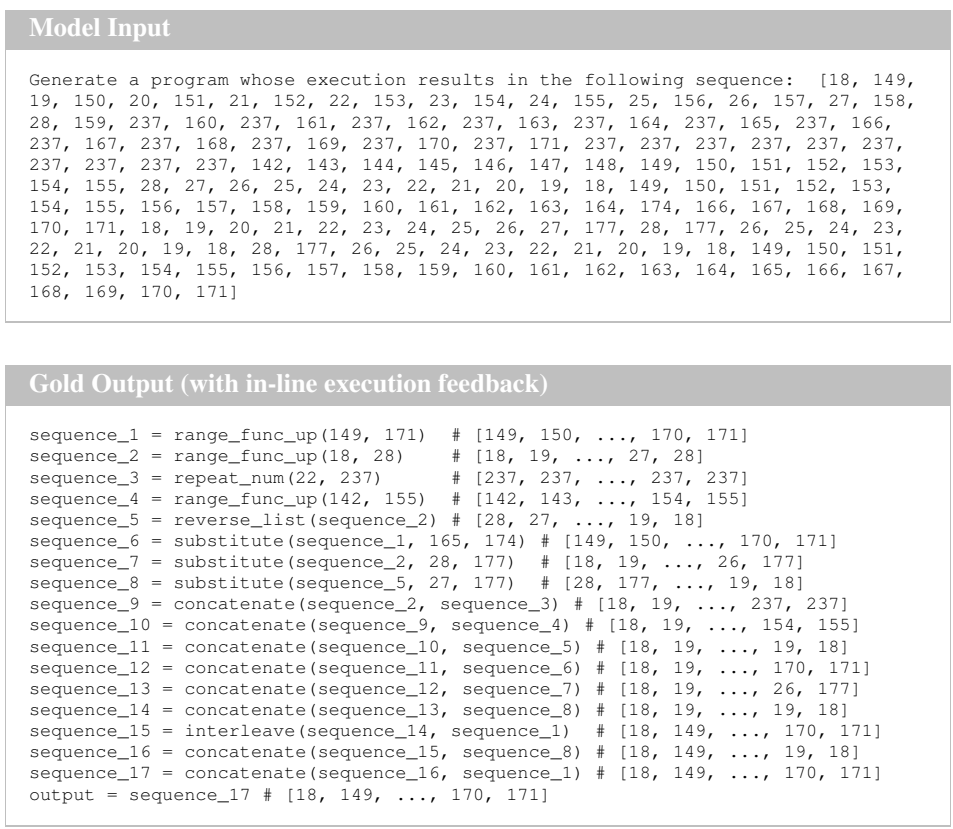

    \centering
    \begin{tcolorbox}[
        colback=white,
        colframe=gray!50,
        title=Model Input,
        width=0.9\textwidth,
        boxrule=0.5pt,
        left=2mm, 
        right=2mm,
        top=2mm,
        bottom=2mm,
        fonttitle=\bfseries,
        sharp corners
        ]
        \ttfamily \scriptsize
        Generate a program whose execution results in the following sequence: 
        [18, 149, 19, 150, 20, 151, 21, 152, 22, 153, 23, 154, 24, 155, 25, 156, 
        26, 157, 27, 158, 28, 159, 237, 160, 237, 161, 237, 162, 237, 163, 237, 
        164, 237, 165, 237, 166, 237, 167, 237, 168, 237, 169, 237, 170, 237, 
        171, 237, 237, 237, 237, 237, 237, 237, 237, 237, 237, 142, 143, 144, 
        145, 146, 147, 148, 149, 150, 151, 152, 153, 154, 155, 28, 27, 26, 25, 
        24, 23, 22, 21, 20, 19, 18, 149, 150, 151, 152, 153, 154, 155, 156, 
        157, 158, 159, 160, 161, 162, 163, 164, 174, 166, 167, 168, 169, 170, 
        171, 18, 19, 20, 21, 22, 23, 24, 25, 26, 27, 177, 28, 177, 26, 25, 
        24, 23, 22, 21, 20, 19, 18, 28, 177, 26, 25, 24, 23, 22, 21, 20, 
        19, 18, 149, 150, 151, 152, 153, 154, 155, 156, 157, 158, 159, 160, 
        161, 162, 163, 164, 165, 166, 167, 168, 169, 170, 171]
    \end{tcolorbox}
    
    \vspace{5pt} 
    
    \begin{tcolorbox}[
        colback=white,
        colframe=gray!50,
        title=Gold Output (with in-line execution feedback),
        width=0.9\textwidth,
        boxrule=0.5pt,
        left=2mm, 
        right=2mm,
        top=2mm,
        bottom=2mm,
        fonttitle=\bfseries,
        sharp corners
        ]
        \ttfamily \scriptsize
\begin{verbatim}
sequence_1 = range_func_up(149, 171)  # [149, 150, ..., 170, 171]
sequence_2 = range_func_up(18, 28)    # [18, 19, ..., 27, 28]
sequence_3 = repeat_num(22, 237)      # [237, 237, ..., 237, 237]
sequence_4 = range_func_up(142, 155)  # [142, 143, ..., 154, 155]
sequence_5 = reverse_list(sequence_2) # [28, 27, ..., 19, 18]
sequence_6 = substitute(sequence_1, 165, 174) # [149, 150, ..., 170, 171]
sequence_7 = substitute(sequence_2, 28, 177)  # [18, 19, ..., 26, 177]
sequence_8 = substitute(sequence_5, 27, 177)  # [28, 177, ..., 19, 18]
sequence_9 = concatenate(sequence_2, sequence_3) # [18, 19, ..., 237, 237]
sequence_10 = concatenate(sequence_9, sequence_4) # [18, 19, ..., 154, 155]
sequence_11 = concatenate(sequence_10, sequence_5) # [18, 19, ..., 19, 18]
sequence_12 = concatenate(sequence_11, sequence_6) # [18, 19, ..., 170, 171]
sequence_13 = concatenate(sequence_12, sequence_7) # [18, 19, ..., 26, 177]
sequence_14 = concatenate(sequence_13, sequence_8) # [18, 19, ..., 19, 18]
sequence_15 = interleave(sequence_14, sequence_1)  # [18, 149, ..., 170, 171]
sequence_16 = concatenate(sequence_15, sequence_8) # [18, 149, ..., 19, 18]
sequence_17 = concatenate(sequence_16, sequence_1) # [18, 149, ..., 170, 171]
output = sequence_17 # [18, 149, ..., 170, 171]
\end{verbatim}
    \end{tcolorbox}

    \caption{An example program-sequence pair with in-line execution feedback.}
    \label{fig:example-ex-feedback}
\end{figure}

\begin{table}[t]
\centering
\tiny
\begin{tabular}{l|cc|cc|cc|cc|cc|cc}
\toprule
           & \multicolumn{2}{c}{Audio-16-bit} & \multicolumn{2}{c}{Audio-8-bit}  & \multicolumn{2}{c}{Audio-MFCC} & \multicolumn{2}{c}{DNA} & \multicolumn{2}{c}{Text} & \multicolumn{2}{c}{Synthetic} \\ 
           & Rep  & No Rep & Rep  & No Rep & Rep  & No Rep & Rep  & No Rep & Rep  & No Rep & Rep  & No Rep \\ \midrule
\llamaxl{} & 84    & 16      & 96    & 4       & 100   & 4       & 100   & 0       & 96    & 4       & 20    & 80     \\ 
\gpt{}     & 96    & 4       & 96    & 4       & 100   & 0       & 100   & 0       & 100   & 0       & 20    & 80     \\ 
\bottomrule
\end{tabular}
\caption{The percentage of repetitions in correct programs from a qualitative analysis of 25 programs per modality for \llamaxl{} and \gpt{}. Most correct examples for natural sequences are because of repetitions, but the contrary is true for synthetic sequences.}
\label{analysis:repetitions}
\end{table}

\begin{table}[ht]
\centering
\tiny
\begin{tabular}{l|ccc|ccc|ccc|ccc|ccc|ccc}
\toprule
           & \multicolumn{3}{c}{16-bit} & \multicolumn{3}{c}{8-bit} & \multicolumn{3}{c}{MFCC} & \multicolumn{3}{c}{DNA} & \multicolumn{3}{c}{Text} & \multicolumn{3}{c}{Synthetic} \\ 
           & Ex  & Re  & Un & Ex  & Re  & Un & Ex  & Re  & Un & Ex  & Re  & Un & Ex  & Re  & Un & Ex  & Re  & Un \\ \midrule
\textsc{LMA-405B} & 80     & 20    & 0     & 96     & 4     & 0     & 72     & 28    & 0     & 40     & 60    & 0     & 24     & 56    & 20    & 92     & 8     & 0     \\ 
\gpt{}     & 72     & 28    & 0     & 88     & 12    & 0     & 40     & 60    & 0     & 36     & 64    & 0     & 44     & 48    & 8     & 96     & 4     & 0     \\ 
\bottomrule
\end{tabular}
\caption{The percentage of execution (Ex), repetition (Re), and Unicode (Un) errors in erroneous programs from a qualitative analysis of 25 programs per modality for \llamaxl{} and \gpt{}. More than a third of the errors for natural sequences are caused by repetition errors.}
\label{analysis:errors}
\end{table}

Fig.~\ref{fig:syn_program_lenghts} and Fig.~\ref{fig:syn_sequence_lenghts} present accuracy for \seqcodertiny{} as a function of program and sequence length, respectively. Accuracy decreases with length of programs or sequences.

Fig.~\ref{fig:example-ex-feedback} presents an example for a synthetic program-sequence pair with in-line execution feedback.

Tab.~\ref{analysis:repetitions} presents the percentage of repetitions in correct programs for \llamaxl{} and \gpt{}. Fig.~\ref{fig:ex-repeat-success} presents an example where \llamaxl{} succeeds by repeating the input sequence. Fig.~\ref{fig:ex-err-repeat} presents an example where \llamaxl{} repeats the input sequence in a complex and verbose manner, which is sub-optimal for the \gzip{} prior. Fig.~\ref{fig:ex-synthetic-success} presents an example where \llamaxl{} succeeds on a synthetic sequence by leveraging the induced patterns.

Tab.~\ref{analysis:errors} presents our qualitative error analysis on 25 erroneous programs per modality for \llamaxl{} and \gpt{}. 
Fig.~\ref{fig:ex-execution-err}, Fig.~\ref{fig:ex-err-syn}, Fig.~\ref{fig:ex-repeating-err}, and Fig.~\ref{fig:ex-text-err} present examples where \llamaxl{} errs due to a wrong execution on an Audio-8-bit, a wrong execution on a synthetic sequence, repetition on an Audio-8-bit sequence, and conversion from text to bytes for a text sequence. 

\begin{figure}[ht]
    \centering
    \scriptsize 
    \begin{tabular}{|p{4cm}|p{5cm}|p{4cm}|}
        \hline
        \multicolumn{1}{|c|}{\textbf{Input Sequence}} & 
        \multicolumn{1}{c|}{\textbf{Generated Program}} & 
        \multicolumn{1}{c|}{\textbf{Produced Sequence}} \\
        \hline
        \begin{flushleft}
            [249, 252, 254, 255, 254, 249, 246, 250, \newline
            254, 255, 255, 253, 252, 254, 255, 0, \newline
            1, 1, 3, 5, 3, 1, 1, 4, 6, 6, 4, 1, 0, \newline
            1, 3, 3, 0, 254, 255, 255, 254, 254, \newline
            253, 253, 254, 254, 253, 253, 253, 255, \newline
            1, 0, 255, 255, 255, 0, 2, 2, 1, 1, 0, \newline
            1, 1, 0, 1, 1, 0, 0, 255, 254, 254, 255, \newline
            255, 254, 253, 252, 254, 254, 255, 255, \newline
            254, 253, 253, 255, 254, 255, 255, 255, \newline
            255, 255, 254, 255, 0, 0, 2, 4, 5, 4, 2, \newline
            1, 2, 5, 6, 6, 3, 1, 0, 1, 2, 1, 1, 0, \newline
            255, 254, 253, 253, 254, 255, 255, 254, \newline
            251, 251, 253, 255, 0, 255, 255, 254, \newline
            254, 255, 1, 2]
        \end{flushleft} &
        \begin{lstlisting}[basicstyle=\ttfamily\scriptsize, breaklines=true]
a = [249, 252, 254, 255]
b = [254, 249, 246, 250, 254, 255, 255, 253, 252, 254, 255]
c = [0, 1, 1, 3, 5, 3, 1, 1, 4, 6, 6, 4, 1, 0, 1, 3, 3, 0]
d = [254, 255, 255, 254, 254, 253, 253, 254, 254, 253, 253, 253, 255]
e = [1, 0, 255, 255, 255, 0, 2, 2, 1, 1, 0, 1, 1, 0, 1, 1, 0, 0]
f = [255, 254, 254, 255, 255, 254, 253, 252, 254, 254, 255, 255, 254]
g = [253, 253, 255, 254, 255, 255, 255, 255, 255, 254, 255]
h = [0, 0, 2, 4, 5, 4, 2, 1, 2, 5, 6, 6, 3, 1, 0, 1, 2, 1, 1, 0]
i = [255, 254, 253, 253, 254, 255, 255, 254, 251, 251, 253, 255]
j = [0, 255, 255, 254, 254, 255, 1, 2]
output = a + b + c + d + e + f + g + h + i + j
        \end{lstlisting} &
        \begin{flushleft}
            [249, 252, 254, 255, 254, 249, 246, 250, \newline
            254, 255, 255, 253, 252, 254, 255, 0, \newline
            1, 1, 3, 5, 3, 1, 1, 4, 6, 6, 4, 1, 0, \newline
            1, 3, 3, 0, 254, 255, 255, 254, 254, \newline
            253, 253, 254, 254, 253, 253, 253, 255, \newline
            1, 0, 255, 255, 255, 0, 2, 2, 1, 1, 0, \newline
            1, 1, 0, 1, 1, 0, 0, 255, 254, 254, 255, \newline
            255, 254, 253, 252, 254, 254, 255, 255, \newline
            254, 253, 253, 255, 254, 255, 255, 255, \newline
            255, 255, 254, 255, 0, 0, 2, 4, 5, 4, 2, \newline
            1, 2, 5, 6, 6, 3, 1, 0, 1, 2, 1, 1, 0, \newline
            255, 254, 253, 253, 254, 255, 255, 254, \newline
            251, 251, 253, 255, 0, 255, 255, 254, \newline
            254, 255, 1, 2]
        \end{flushleft} \\
        \hline
    \end{tabular}
    \caption{An example generation for \llamaxl{} on Audio-8-bit where the model succeeds by splitting the sequence to sub-sequences that are later concatenated.}
    \label{fig:ex-repeat-success}
\end{figure}

\begin{figure}[ht]
    \centering
    \scriptsize 
    \begin{tabular}{|p{4cm}|p{5cm}|p{4cm}|}
        \hline
        \multicolumn{1}{|c|}{\textbf{Input Sequence}} & 
        \multicolumn{1}{c|}{\textbf{Generated Program}} & 
        \multicolumn{1}{c|}{\textbf{Produced Sequence}} \\
        \hline
        \begin{flushleft}
            [250, 250, 251, 251, 251, 250, 248, 246, \newline
            244, 244, 243, 244, 245, 247, 249, 251, \newline
            251, 250, 248, 246, 243, 242, 241, 241, \newline
            241, 241, 242, 243, 244, 245, 245, 246, \newline
            246, 245, 245, 244, 244, 245, 245, 245, \newline
            246, 245, 246, 245, 244, 243, 243, 244, \newline
            244, 245, 246, 247, 249, 255, 16, 34, \newline
            43, 40, 29, 27, 29, 23, 9, 252, 248, 250, \newline
            248, 245, 250, 4, 11, 7, 2, 4, 12, 14, \newline
            9, 4, 8, 16, 21, 22, 26, 31, 30, 22, \newline
            12, 7, 4, 252, 241, 234, 235, 239, 241, \newline
            243, 249, 255, 2, 1, 1, 4, 7, 6, 4, \newline
            5, 9, 13, 13, 11, 9, 6, 1, 249, 243, \newline
            241, 239, 236, 235, 237, 242, 248, 252, \newline
            255, 2, 4, 5, 5]
        \end{flushleft} &
        \begin{lstlisting}[basicstyle=\ttfamily\scriptsize, breaklines=true]
a = 250
b = [a, a]
c = [251, 251, 251]
b.extend(c)
d = [a, 248, 246, 244, 244, 243, 244, 245, 247, 249]
b.extend(d)
e = [251, 251, a, 248, 246, 243, 242, 241, 241, 241, 
     241, 242, 243, 244, 245, 245, 246, 246, 245, 
     245, 244, 244, 245, 245, 245, 246, 245, 246, 
     245, 244, 243, 243, 244, 244, 245, 246, 247, 
     249, 255, 16, 34, 43, 40, 29, 27, 29, 23, 9]
b.extend(e)
f = [252, 248, 250, 248, 245, 250, 4, 11, 7, 2, 4, 
     12, 14, 9, 4, 8, 16, 21, 22, 26, 31, 30, 22, 
     12, 7, 4]
b.extend(f)
g = [252, 241, 234, 235, 239, 241, 243, 249, 255, 2, 
     1, 1, 4, 7, 6, 4, 5, 9, 13, 13, 11, 9, 6, 1]
b.extend(g)
h = [249, 243, 241, 239, 236, 235, 237, 242, 248, 
     252, 255, 2, 4, 5, 5]
b.extend(h)
output = b
        \end{lstlisting} &
        \begin{flushleft}
            [250, 250, 251, 251, 251, 250, 248, 246, \newline
            244, 244, 243, 244, 245, 247, 249, 251, \newline
            251, 250, 248, 246, 243, 242, 241, 241, \newline
            241, 241, 242, 243, 244, 245, 245, 246, \newline
            246, 245, 245, 244, 244, 245, 245, 245, \newline
            246, 245, 246, 245, 244, 243, 243, 244, \newline
            244, 245, 246, 247, 249, 255, 16, 34, \newline
            43, 40, 29, 27, 29, 23, 9, 252, 248, 250, \newline
            248, 245, 250, 4, 11, 7, 2, 4, 12, 14, \newline
            9, 4, 8, 16, 21, 22, 26, 31, 30, 22, \newline
            12, 7, 4, 252, 241, 234, 235, 239, 241, \newline
            243, 249, 255, 2, 1, 1, 4, 7, 6, 4, \newline
            5, 9, 13, 13, 11, 9, 6, 1, 249, 243, \newline
            241, 239, 236, 235, 237, 242, 248, 252, \newline
            255, 2, 4, 5, 5]
        \end{flushleft} \\
        \hline
    \end{tabular}
    \caption{An example generation from \llamaxl{} for Audio-8-bit where the program repeats the input sequence in a verbose way, i.e., by extracting the number $255$ to a variable and splitting to sub-sequences.}
    \label{fig:ex-err-repeat}
\end{figure}

\begin{figure}[ht]
    \centering
    \scriptsize 
    \begin{tabular}{|p{4cm}|p{5cm}|p{4cm}|}
        \hline
        \multicolumn{1}{|c|}{\textbf{Input Sequence}} & 
        \multicolumn{1}{c|}{\textbf{Generated Program}} & 
        \multicolumn{1}{c|}{\textbf{Produced Sequence}} \\
        \hline
        \begin{flushleft}
            [196, 197, 198, 199, 200, 201, 202, 203, \newline
            204, 205, 206, 207, 208, 209, 210, 211, \newline
            212, 213, 80, 80, 80, 80, 80, 80, 80, 26, \newline
            26, 26, 26, 26, 108, 97, 19, 251, 31, 45, \newline
            93, 136, 249, 32, 162, 7, 184, 244, 191, \newline
            178, 179, 180, 181, 182, 183, 184, 185, \newline
            186, 187, 188, 189, 190, 191, 192, 193, \newline
            194, 195, 196, 197, 198, 199, 200, 201, \newline
            80, 80, 80, 80, 80, 80, 80]
        \end{flushleft} &
        \begin{lstlisting}[basicstyle=\ttfamily\scriptsize, breaklines=true]
output = (
    list(range(196, 214)) 
    + [80] * 7 
    + [26] * 5 
    + [108, 97, 19, 251, 31, 45, 93, 136, 249, 32, 162, 7, 184, 244, 191] 
    + list(range(178, 202)) 
    + [80] * 7
)
        \end{lstlisting} &
        \begin{flushleft}
            [196, 197, 198, 199, 200, 201, 202, 203, \newline
            204, 205, 206, 207, 208, 209, 210, 211, \newline
            212, 213, 80, 80, 80, 80, 80, 80, 80, 26, \newline
            26, 26, 26, 26, 108, 97, 19, 251, 31, 45, \newline
            93, 136, 249, 32, 162, 7, 184, 244, 191, \newline
            178, 179, 180, 181, 182, 183, 184, 185, \newline
            186, 187, 188, 189, 190, 191, 192, 193, \newline
            194, 195, 196, 197, 198, 199, 200, 201, \newline
            80, 80, 80, 80, 80, 80, 80]
        \end{flushleft} \\
        \hline
    \end{tabular}
    \caption{An example synthetic sequence where \llamaxl{} succeeds in leveraging patterns in the input sequence, i.e., repetitions and an increasing range, to generate a correct program.}
    \label{fig:ex-synthetic-success}
\end{figure}

\begin{figure}[ht]
    \centering
    \scriptsize 
    \begin{tabular}{|p{4cm}|p{5cm}|p{4cm}|}
        \hline
        \multicolumn{1}{|c|}{\textbf{Input Sequence}} & 
        \multicolumn{1}{c|}{\textbf{Generated Program}} & 
        \multicolumn{1}{c|}{\textbf{Produced Sequence}} \\
        \hline
        \begin{flushleft}
            [6, 7, 7, 8, 10, 11, 11, 11, \newline
            7, 3, 3, 6, 4, 254, 247, 246, \newline
            248, 252, 252, 248, 245, 246, \newline
            250, 254, 254, 253, 253, 254, \newline
            255, 0, 255, 255, 255, 254, \newline
            254, 254, 253, 253, 254, 255, \newline
            254, 252, 252, 253, 254, 255, \newline
            254, 253, 253, 254, 0, 0, \newline
            255, 254, 253, 254, 254, 254, \newline
            253, 253, 253, 254, 254, 254, \newline
            255, 0, 2, 2, 2, 1, 2, 3, \newline
            4, 3, 2, 1, 1, 1, 1, 0, \newline
            255, 253, 253, 253, 253, 253, \newline
            253, 253, 254, 255, 0, 0, \newline
            1, 3, 4, 5, 6, 7, 9, 10, \newline
            11, 12, 9, 6, 5, 7, 7, 2, \newline
            252, 249, 249, 251, 252, 249, \newline
            246, 245, 247, 251, 252, 252, \newline
            252, 253, 254, 255, 255, 0, 0, 0]
        \end{flushleft} &
        \begin{lstlisting}[basicstyle=\ttfamily\scriptsize, breaklines=true]
# Predefined Constants
start_sequence = [6, 7, 7, 8, 10, 11, 11, 11, 7, 3, 3, 6, 4]
mid_sequence = [255, 0, 255, 255, 255]
end_sequence = [0, 0, 0]

# Pattern 1: Decreasing by 1
def decreasing_pattern(start, length):
    return [start - i for i in range(length)]

# Pattern 2: Increasing by 1
def increasing_pattern(start, length):
    return [start + i for i in range(length)]

# Pattern 3: Repeating a number
def repeat_pattern(num, times):
    return [num] * times

# Generate the Sequence
sequence = start_sequence
sequence += decreasing_pattern(254, 5)
sequence += increasing_pattern(245, 5)
sequence += repeat_pattern(254, 3)
sequence += [253, 253, 254, 255]
sequence += mid_sequence
sequence += repeat_pattern(254, 3)
sequence += decreasing_pattern(254, 3)
sequence += [0, 0, 255]
sequence += repeat_pattern(254, 3)
sequence += decreasing_pattern(254, 4)
sequence += [253, 253, 253, 254, 254, 254, 255, 0]
sequence += [2, 2, 2, 1, 2, 3, 4, 3, 2, 1, 1, 1, 1, 0]
sequence += mid_sequence
sequence += repeat_pattern(253, 7)
sequence += [254, 255, 0, 0, 1, 3, 4, 5, 6, 7, 9, 10, 11, 12, 9, 6, 5, 7, 7, 2]
sequence += decreasing_pattern(252, 3)
sequence += [249, 246, 245, 247, 251, 252, 252, 252, 253, 254, 255, 255]
sequence += end_sequence

# Final Output
output = sequence
        \end{lstlisting} &
        \begin{flushleft}
            [6, 7, 7, 8, 10, 11, 11, 11, \newline
            7, 3, 3, 6, 4, 254, 253, 252, \newline
            251, 250, 245, 246, 247, 248, \newline
            249, 254, 254, 254, 253, 253, \newline
            254, 255, 255, 0, 255, 255, \newline
            255, 254, 254, 254, 254, 253, \newline
            252, 0, 0, 255, 254, 254, \newline
            254, 254, 253, 252, 251, 253, \newline
            253, 253, 254, 254, 254, 255, \newline
            0, 2, 2, 2, 1, 2, 3, 4, \newline
            3, 2, 1, 1, 1, 1, 0, 255, \newline
            0, 255, 255, 255, 253, 253, \newline
            253, 253, 253, 253, 253, 254, \newline
            255, 0, 0, 1, 3, 4, 5, 6, \newline
            7, 9, 10, 11, 12, 9, 6, 5, \newline
            7, 7, 2, 252, 251, 250, 249, \newline
            246, 245, 247, 251, 252, 252, \newline
            252, 253, 254, 255, 255, 0, 0, 0]
        \end{flushleft} \\
        \hline
    \end{tabular}
    \caption{An example program for Audio-8-bit from \llamaxl{} where the model tries to identify patterns in the input sequence, but the execution does not result in the input.}
    \label{fig:ex-execution-err}
\end{figure}

\begin{figure}[ht]
    \centering
    \scriptsize 
    \begin{tabular}{|p{4cm}|p{5cm}|p{4cm}|} 
        \hline
        \multicolumn{1}{|c|}{\textbf{Input Sequence}} & 
        \multicolumn{1}{c|}{\textbf{Generated Program}} & 
        \multicolumn{1}{c|}{\textbf{Produced Sequence}} \\
        \hline
        \begin{flushleft}
            [140, 55, 142, 57, 144, 59, 146, 61, \newline
            148, 63, 150, 65, 152, 67, 154, 69, \newline
            156, 71, 1, 73, 77, 75, 101, 77, 45, \newline
            79, 160, 81, 29, 83, 187, 85, 236, \newline
            87, 65, 89, 196, 91, 18, 93, 68, 95, \newline
            139, 97, 221, 99, 165, 166, 167, 168, \newline
            169, 170, 171, 172, 173, 174, 175, \newline
            176, 55, 57, 59, 61, 63, 65, 67, 69, \newline
            71, 73, 75, 77, 79, 81, 83, 85, 87, \newline
            89, 91, 93, 95, 97, 99]
        \end{flushleft} &
        \begin{lstlisting}[basicstyle=\ttfamily\scriptsize, breaklines=true]
a = 140
b = 55
c = 2
d = [a + i * c for i in range(10)]
e = [b + i * c for i in range(20)]
f = [101, 45, 160, 29, 187, 236, 65,
     196, 18, 68, 139, 221, 165]
g = list(range(166, 177))
h = [55 + i * 2 for i in range(23)]
output = d[0:10] + e[0:10] + [1] + e[10:12]
      + f + e[12:15] + g + h
        \end{lstlisting} &
        \begin{flushleft}
            [140, 142, 144, 146, 148, 150, 152, \newline
            154, 156, 158, 55, 57, 59, 61, 63, 65, \newline
            67, 69, 71, 73, 1, 75, 77, 101, 45, 160, \newline
            29, 187, 236, 65, 196, 18, 68, 139, 221, \newline
            165, 79, 81, 83, 166, 167, 168, 169, 170, \newline
            171, 172, 173, 174, 175, 176, 55, 57, 59, \newline
            61, 63, 65, 67, 69, 71, 73, 75, 77, 79, \newline
            81, 83, 85, 87, 89, 91, 93, 95, 97, 99]
        \end{flushleft} \\
        \hline
    \end{tabular}
    \caption{An example program for a synthetic sequence from \llamaxl{} where the model tries to identify patterns in the input sequence, but the execution does not result in the input.}
    \label{fig:ex-err-syn}
\end{figure}

\begin{figure}[ht]
    \centering
    \scriptsize 
    \begin{tabular}{|p{4cm}|p{5cm}|p{4cm}|}
        \hline
        \multicolumn{1}{|c|}{\textbf{Input Sequence}} & 
        \multicolumn{1}{c|}{\textbf{Generated Program}} & 
        \multicolumn{1}{c|}{\textbf{Produced Sequence}} \\
        \hline
        \begin{flushleft}
            [235, 236, 240, 244, 248, 252, 255, 1, \newline
            4, 6, 8, 10, 10, 12, 14, 16, \newline
            18, 16, 13, 9, 5, 4, 1, 254, \newline
            248, 244, 241, 240, 239, 239, 238, 236, \newline
            234, 233, 234, 236, 237, 237, 237, 238, \newline
            241, 244, 246, 246, 246, 245, 245, 246, \newline
            247, 248, 247, 246, 246, 246, 247, 247, \newline
            246, 245, 243, 243, 244, 245, 246, 246, \newline
            245, 245, 245, 247, 249, 249, 249, 250, \newline
            252, 254, 2, 5, 7, 9, 11, 14, \newline
            18, 22, 27, 30, 34, 37, 39, 40, \newline
            41, 41, 40, 36, 28, 20, 17, 14, \newline
            10, 5, 1, 255, 252, 247, 244, 243, \newline
            240, 235, 233, 234, 238, 242, 246, 249, \newline
            253, 0, 3, 6, 8, 9, 10, 11, 12, 15, \newline
            19, 20, 19, 15, 12]
        \end{flushleft} &
        \begin{lstlisting}[basicstyle=\ttfamily\scriptsize, breaklines=true]
a = [235, 236, 240, 244, 248, 252, 255]
b = [1, 4, 6, 8, 10]
c = [10, 12, 14, 16, 18]
d = [16, 13, 9, 5, 4, 1]
e = [254, 248, 244, 241, 240, 239]
f = [239, 238, 236, 234, 233]
g = [234, 236, 237]
h = [238, 241, 244, 246]
i = [245, 245, 246, 247, 248, 247, 246]
j = [243, 243, 244, 245, 246]
k = [249, 249, 249, 250, 252, 254]
l = [2, 5, 7, 9, 11, 14, 18, 22, 27, 30, 34, 37, 39, 40]
m = [41, 41, 40, 36, 28, 20, 17, 14, 10, 5, 1]
n = [255, 252, 247, 244, 243, 240, 235]
o = [233, 234, 238, 242, 246, 249, 253]
p = [0, 3, 6, 8, 9, 10, 11, 12, 15, 19, 20]
q = [19, 15, 12]

output = a + b + c + c + d + e + f + g + g + g + h + i + i + j + j + k + l + m + n + o + p + q
        \end{lstlisting} &
        \begin{flushleft}
            [235, 236, 240, 244, 248, 252, 255, 1, \newline
            4, 6, 8, 10, 10, 12, 14, 16, \newline
            18, 10, 12, 14, 16, 18, 16, 13, \newline
            9, 5, 4, 1, 254, 248, 244, 241, \newline
            240, 239, 239, 238, 236, 234, 233, 234, \newline
            236, 237, 234, 236, 237, 234, 236, 237, \newline
            238, 241, 244, 246, 245, 245, 246, 247, \newline
            248, 247, 246, 245, 245, 246, 247, 248, \newline
            247, 246, 243, 243, 244, 245, 246, 243, \newline
            243, 244, 245, 246, 249, 249, 249, 250, \newline
            252, 254, 2, 5, 7, 9, 11, 14, \newline
            18, 22, 27, 30, 34, 37, 39, 40, \newline
            41, 41, 40, 36, 28, 20, 17, 14, \newline
            10, 5, 1, 255, 252, 247, 244, 243, \newline
            240, 235, 233, 234, 238, 242, 246, 249, \newline
            253, 0, 3, 6, 8, 9, 10, 11, 12, 15, \newline
            19, 20, 19, 15, 12]
        \end{flushleft} \\
        \hline
    \end{tabular}
    \caption{An example program for Audio-8-bit from \llamaxl{} where the model tries to repeat the input sequence by breaking it to sub-sequences that are later concatenated but still errs.}
    \label{fig:ex-repeating-err}
\end{figure}

\begin{figure}[ht]
    \centering
    \scriptsize 
    \begin{tabular}{|p{4cm}|p{5cm}|p{4cm}|}
        \hline
        \multicolumn{1}{|c|}{\textbf{Input Sequence}} & 
        \multicolumn{1}{c|}{\textbf{Generated Program}} & 
        \multicolumn{1}{c|}{\textbf{Produced Sequence}} \\
        \hline
        \begin{flushleft}
            [32, 102, 111, 114, 99, 101, 115, 32, \newline
            105, 110, 32, 116, 104, 101, 32, 69, \newline
            97, 115, 116, 32, 116, 111, 32, 100, \newline
            101, 97, 108, 32, 119, 105, 116, 104, \newline
            46, 32, 87, 101, 101, 107, 115, 32, \newline
            108, 97, 116, 101, 114, 32, 74, 111, \newline
            104, 110, 115, 116, 111, 110, 32, 119, \newline
            111, 117, 108, 100, 32, 100, 101, 102, \newline
            121, 32, 74, 101, 102, 102, 101, 114, \newline
            115, 111, 110, 32, 68, 97, 118, 105, \newline
            115, 32, 97, 110, 100, 32, 115, 117, \newline
            114, 114, 101, 110, 100, 101, 114, 32, \newline
            104, 105, 115, 32, 102, 111, 114, 99, \newline
            101, 115, 32, 116, 111, 32, 83, 104, \newline
            101, 114, 109, 97, 110, 46, 32, 79, \newline
            102, 32, 99, 111, 117, 114, 115, 101]
        \end{flushleft} &
        \begin{lstlisting}[basicstyle=\ttfamily\scriptsize, breaklines=true]
letters = 'abcdefghijklmnopqrstuvwxyz'
numbers = '0123456789'
sentence = "Forces in the East to deal with. Weeks later Johnston would defy Jefferson Davis and surrender his forces to Sherman. Of course"

output = [ord(c.lower()) if c.lower() in letters else ord(c) for c in sentence]
        \end{lstlisting} &
        \begin{flushleft}
            [102, 111, 114, 99, 101, 115, 32, 105, \newline
            110, 32, 116, 104, 101, 32, 101, 97, \newline
            115, 116, 32, 116, 111, 32, 100, 101, \newline
            97, 108, 32, 119, 105, 116, 104, 46, \newline
            32, 119, 101, 101, 107, 115, 32, 108, \newline
            97, 116, 101, 114, 32, 106, 111, 104, \newline
            110, 115, 116, 111, 110, 32, 119, 111, \newline
            117, 108, 100, 32, 100, 101, 102, 121, \newline
            32, 106, 101, 102, 102, 101, 114, 115, \newline
            111, 110, 32, 100, 97, 118, 105, 115, \newline
            32, 97, 110, 100, 32, 115, 117, 114, \newline
            114, 101, 110, 100, 101, 114, 32, 104, \newline
            105, 115, 32, 102, 111, 114, 99, 101, \newline
            115, 32, 116, 111, 32, 115, 104, 101, \newline
            114, 109, 97, 110, 46, 32, 111, 102, \newline
            32, 99, 111, 117, 114, 115, 101]
        \end{flushleft} \\
        \hline
    \end{tabular}
    \caption{An example program for the text modality from \llamaxl{} where the model tries to produce the sequence by writing a text and converting it to Unicode bytes, but errs slightly (the correct text being \texttt{" forces in the East to deal with.~Weeks later Johnston would defy Jefferson Davis and surrender his forces to Sherman.~Of course"}).}
    \label{fig:ex-text-err}
\end{figure}

\end{document}